\documentclass[draftcls,12pt,onecolumn]{IEEEtran} 
\usepackage{multirow}
\usepackage{amsfonts}
\usepackage{amssymb}
\usepackage{amsbsy} 
\usepackage{amsmath}
\usepackage[ansinew]{inputenc}%
\usepackage{cite}
\usepackage{pstricks}
\usepackage{multirow}
\usepackage{color}
\usepackage{lscape}
\usepackage[dvips]{graphicx}
\usepackage{colortbl}
\usepackage{url}
\usepackage{subfigure}
\usepackage{hyperref}

\usepackage{algorithmic}
\usepackage{algorithm}

\renewcommand{\white}[1]{\textcolor[rgb]{1,1,1}{#1}}

\renewcommand{\blue}[1]{\textcolor[rgb]{0,0,0}{#1}}

\newcommand{\mytitle}{Learning user's confidence for active learning}

\def\x{{\mathbf x}}

\begin{document}

\title{\mytitle}
\author{Devis Tuia,~\IEEEmembership{Member,~IEEE},
Jordi Mu{\~n}oz-Mar{\'i}~\IEEEmembership{Member,~IEEE}
\thanks{Manuscript received XXXX;}
\thanks{This work has been partly supported by the Swiss National
Science Foundation (grant PZ00P2-136827) and by the Spanish Ministry of Science and Innovation under the projects CICYT-FEDER TEC2009-13696, AYA2008-05965-C04-03, and CSD2007-00018.}
\thanks{ DT is with the Laboratoire des Syst\`emes d'Information G\'eographique (LaSIG), Lausanne Institute of Technology (EPFL), Switzerland. devis.tuia@epfl.ch, http://devis.tuia.googlepages.com, Phone: +41-216935785, Fax : +41-216935790.
\newline JMM is with the Image Processing Laboratory (IPL),
Universitat de Val{\`{e}}ncia, Val{\`{e}}ncia, Spain.
E-mail: jordi.munoz@uv.es, http://isp.uv.es, Phone: +34-963544021, Fax: +34-963544353.
}}

\markboth{IEEE Transactions on Geoscience and Remote Sensing, preprint. Published version (2013): 10.1109/TGRS.2012.2203605}{Tuia et al.: \mytitle}

\maketitle


\begin{abstract}
\textbf{This is the pre-acceptance version, to read the final version published in 2013 in the IEEE Transactions on Geoscience and Remote Sensing (IEEE TGRS), please go to: \href{https://doi.org/10.1109/TGRS.2012.2203605}{10.1109/TGRS.2012.2203605}}\\
In this paper, we study the applicability of active learning in operative scenarios: more particularly, we consider the well-known contradiction between the active learning heuristics, which rank the pixels according to their uncertainty, and the user's confidence in labeling, which is related to both the homogeneity of the pixel context and user's knowledge of  the scene. We propose a filtering scheme based on a classifier that learns the confidence of the user in labeling, thus minimizing the queries where the user would not be able to provide a class for the pixel. The capacity of a model to learn the user's confidence is studied in detail, also showing the effect of resolution is such a learning task. Experiments on two QuickBird images of different resolutions (with and without pansharpening) and considering committees of users prove the efficiency of the filtering scheme proposed, which maximizes the number of useful queries with respect to traditional active learning.
\end{abstract}


\section{Introduction}
\label{sec:intro}

\PARstart{T}{he} advent of remote sensing imagery has opened a wide range of possibilities for surveying and analyzing the processes occurring on the surface of the Earth, thus allowing significant advances in the monitoring of agricultural~\cite{Pin03b,Dor07} or urban processes~\cite{Tau12}. Among all the products retrieved from very high resolution (VHR) imagery, classification maps describing landuse remain the most common. Several methods have been proposed to perform the classification task, but until now supervised methods remain the most successful approaches in the remote sensing community~\cite{Cam11,Pra11}. However, these approaches rely on a set of pixels for which the class is known and that are used to train the classifier: the training set. The representativity of this set is crucial for the success of the classification~\cite{Foo04,Foo06}. There are two major ways of obtaining a training set: one is to organize in-situ campaigns, where the landuse represented by the pixel is assessed and georeferenced by teams on the field; the other is to proceed by photointerpretation, i.e., having a human operator define labeled polygons on screen. Photointerpretation is particularly successful when dealing with VHR images, since the objects are recognizable on screen. In this case, two problems arise: first, many redundant pixels are added to the training set and second, the pixels added are not necessarily the most relevant for the classifier, but those that were most convenient (for a variety of reasons) in the photointerpretation phase. This last point is crucial, since the photointerpreter tends to label easily-recognizable pixels and to avoid areas of high variance/contrast (unless this variance is the specificity of the class) or underrepresented landuse classes.

To make photointerpretation efficient, active learning methods have recently been proposed in the community (a review in~\cite{Tui11}): with active learning, the model and the user interact, the first ranking unlabeled pixels by their classification uncertainty and the latter providing the labels of the highly ranked. After retraining with these difficult pixels (now labeled), the model is expected to  improve its performance greatly.

Several ranking criteria (heuristics) have been proposed in the remote sensing literature: some use committees of machines working on subsets of training pixels~\cite{Tui09} or of input features~\cite{Di10c,Di12}, others use the SVM decision function~\cite{Mit04,Pas10c,Pat10}, posterior probabilities~\cite{Li10b,Tui11d,Li11} or cluster coherence~\cite{Tui10d,Mun11} as a criterion to rank pixels. Questions of batch diversity~\cite{Tui09,Dem10}, inter-iterations diversity~\cite{Vol10d} and inter-dataset adaptation~\cite{Tui11d,Persello2011} have also been considered. All these studied  proved the efficiency of active learning heuristics in querying the most informative and diverse pixels in a pool of possible candidates.

Despite the theoretical appeal of this solution, the constraints of photointerpretation are often contradictory to the common active learning setting: while the first are driven by the user's capacity to recognize the objects on the surface, the second ranks the pixels by their uncertainty, i.e., the complexity and mixture of their signature. As a consequence, the user is constantly required to label those samples with the highest uncertainty, which is a very complex (and often unfeasible) task even for a trained operator.

\begin{figure}[t]
\centering
\begin{tabular}{ccc}
\includegraphics[width = 2.5cm,height = 2.5cm]{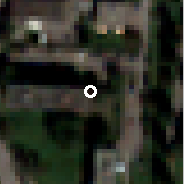}& 
\includegraphics[width = 2.5cm,height = 2.5cm]{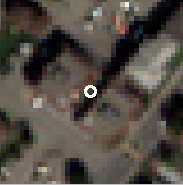}& 
\includegraphics[width = 2.5cm,height = 2.5cm]{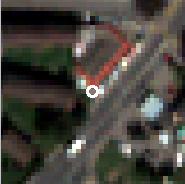}\\
\includegraphics[width = 2.5cm,height = 2.5cm]{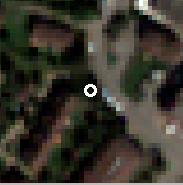}& 
\includegraphics[width = 2.5cm,height = 2.5cm]{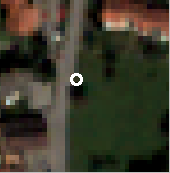}& 
\includegraphics[width = 2.5cm,height = 2.5cm]{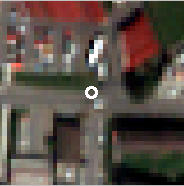}\\
\end{tabular}
\caption{Six examples of pixels (in the white circles) selected to be labeled by a standard active learning criterion. Most of them are placed in areas under shadows, or at the boundary between several classes.}
\label{fig:pix}
\end{figure}

Figure~\ref{fig:pix} illustrates this principle for a 2.4~m QuickBird image: frequently, the pixels queried by an active learning heuristic are situated on the borders between objets, in areas which are not homogeneous, between several classes or in shadowed areas.

In all recent active learning literature (see above), this problem was avoided since, in order to avoid tedious photointerpretation, a pre-labeled set of candidates was provided and the human user was replaced by the a-priori known labels. This produces two undesired effects: first, all the borders between the objects were avoided, since the pre-labeled set is usually defined by photointerpretation and does not cover the whole of the image; and secondly, the user was considered infallible, in the sense that he could always give the correct answer. Reality, however, is often very different, and even an experienced operator can have difficulties in labeling the pixels returned by an active learning heuristic.

In this paper, we propose a solution to maximize the chances of querying pixels that the user can label correctly. To do so, we consider the idea of \emph{learning the confidence of the user in a photointerpreting task}. Each time a user is unable to label a query, he gives information about a query he does not want to encounter again in the process, i.e., a bad state~\cite{Jud11}. Avoiding bad states through estimation of the confidence of the oracle performing the task has been studied in robotics~\cite{Ina99,Nic03,Gro07} and for time-evolving queries in policy evaluation~\cite{Che09}. 

To avoid bad states, we propose to train a second classifier learning to separate valid states (pixels that the user can label) from bad states. Using the confidence of this classifier, we build a mask for the active learning heuristic, which then avoids queries where the model learnt that the user was unable to give an answer. In short, we introduce the concept of learning the confidence of the user to avoid bad states when performing active learning while he/she is photointerpreting. We apply the proposed technique to two VHR images of urban areas. In both cases, the proposed active learning system allows to reconstruct the scene with a minimal number of queries, also minimizing the number of bad states. This means that for the same effort on the part of the user, the learning curve is steeper and all the queries presented are answered and are relevant for the model. Finally, the proposed algorithm also returns a confidence map that shows where the user would be capable to provide an answer.

The remainder of the paper is organized as follows: Section~\ref{sec:badstate} illustrates the proposed methodology. Section~\ref{sec:data} describes the data considered and the experimental setup of the experiments. Section~\ref{sec:res} presents the results ans Section~\ref{sec:concl} gives the conclusions of the work.

\section{Active learning with user's confidence}~\label{sec:badstate}

This section presents the Active Learning with User's Confidence (AL-UC) method proposed. We first review the standard active learning framework and then present the strategy used to avoid bad states.

\subsection{Active learning and uncertainty sampling}
Active learning~\cite{Coh94,Set10} is a way to address the problem of ranking a set of unlabeled pixels, $U$, according to a score providing information about their classification uncertainty. Starting at iteration $\epsilon = 1$ with a training set composed by $l$ labeled pixels, $X^\epsilon = \{\x_i,y_i\}_{i = 1}^l$, a supervised model is trained and the pixels in $U^\epsilon$ are ranked according to a heuristic accounting for the content in information carried by every unlabeled pixels for the model at the current iteration $\epsilon$. The pixels related to maximal uncertainty are presented to an oracle (a photointerpreter in our case) who labels them, thus discovering their labels. The $m$ newly labeled pixels form a batch $S^\epsilon = \{\x_j,y_j\}_{j = 1}^m$ that is added to the training set ($X^{\epsilon+1} = X^\epsilon \cup S^\epsilon$) and removed from the unlabeled set of candidates ($U^{\epsilon+1} = U \setminus S^\epsilon$). As stated in the introduction, many heuristics exist to perform the ranking of the candidates. In this paper, we used \blue{two different heuristics:}

\begin{itemize}
\item[-] The Multi-Class Level Uncertainty criterion proposed in~\cite{Dem10}, which is a state of the art criterion for uncertainty sampling. This criterion, based on the SVM decision function~\cite{Bos92}, ranks the pixels by confronting output of the two most confident classes in a One-Against-All setting. For a given iteration $t$ and $\Omega$ possible classes, the most uncertain pixel is the one for which

\begin{align}
&\hat{\x}^{\text{MCLU}} = \arg \min_{\x_i \in U} \Big\{f(\x_i)^{\text{MC}}\Big\}\label{eq:MCLU}\\
&\text{where} \qquad f(\x_i)^{\text{MC}} = \max_{\omega \in \Omega}|f(\x_i,\omega)|-\max_{\omega \in \Omega\backslash \omega^+}|f(\x_i,\omega)|\label{eq:MCLUfct}
\end{align}

where $\omega^+$ is the class showing maximal confidence, i.e. the argument of the first term of Eq.~\eqref{eq:MCLUfct}. A high value of this criterion corresponds to samples assigned with high certainty to the most confident class, while a small value represents unreliable classification.

\item[-] \blue{The Entropy-Query-by-Bagging (EQB) proposed in~\cite{Tui09}. This criterion ranks the candidates using a committee of classifiers trained with a subset of the training data $X^\epsilon$. Pixels related to maximal entropy in the predictions given by the committee ($H^{\text{BAG}}$) are retained.  In this paper, we use the normalized version of the heuristic ($nEQB$)~\cite{Tui11}}

\begin{equation}
	\hat{\x}^{\text{\emph{n}EQB}} = \arg\max_{\x_i \in U}\Big\{\frac{H^{\text{BAG}}(\x_i)}{ \mbox{log}(N_i)}\Big\}
\label{eq:neqb}
\end{equation}

where

\begin{align}
	&H^{\text{BAG}}(\x_i) = -\sum_{\omega=1}^{N_i} p^{\text{BAG}}(y_i^* = \omega|\x_i) \mbox{log}\left[p^{\text{BAG}}(y^*_i= \omega|\x_i)\right]\\
		& \text{where} \qquad p^{\text{BAG}}(y^*_i = \omega|\x_i) = \frac{\sum_{m=1}^k \delta(y_{i,m}^{*}, \omega )}{\sum_{m=1}^k\sum_{j = 1}^{N_i} \delta(y_{i,m}^*, \omega_j ) } \nonumber 
\label{eq:entropy}
\end{align}

\blue{$N_i$ is the number of classes predicted for pixel $\x_i$ by the committee, with $1 \leq N_i \leq N$. $N$ is the total number of classes. $\delta$ is a function returning $1$ if the class predicted  by the $m$-th member of the committee is $\omega_j$ and $0$ otherwise.}

\end{itemize}

Once the ranking of the $U$ set is provided, the user is asked to label the pixels minimizing the criterion in~\eqref{eq:MCLU} \blue{or \eqref{eq:neqb} respectively}. \blue{This paper only considers these two heuristics because} i) all heuristics based on pixel's uncertainty only produce similar results~\cite{Tui11} and ii) including a criterion on pixel diversity would make the queries more effective~\cite{Dem10}, but do not change the conclusions in the optic of this study. \blue{Therefore, we limit the experimentation  to these two heuristics.}

\subsection{Learning user's confidence}
Assuming that the photointerpreter is able to label all the pixels in $U$ makes active learning algorithms efficient tools for semiautomatic training set composition. However, the pixels minimizing Eq.~\eqref{eq:MCLU} are often difficult to label, since  i) they correspond to pixels related to maximal uncertainty ii) they are situated on the border between classes. As a consequence, these pixels often lie on the border between objects in the spatial domain, as observed in the zooms reported in Fig.~\ref{fig:pix}.

These difficult pixels can be considered as bad states, in the sense that the user may be uncertain or agnostic about the response to give for these queries, since their choice does not depend on the ability of the user nor on his/her knowledge of the scene~\cite{Jud11}. 

A user can become frustrated when encountering several bad states, since he/she is forced to use useless resources 
(he / she cannot provide an answer repeatedly, and the time needed to get the necessary number of labeled pixels is therefore greatly lengthened) at the risk of degrading his/her performances (for example for an increased fatigue). Worse, the user can decide to give advice that is unreliable, thus degrading the model performance with mislabeled training samples.

To decide which states the user would or wouldn't like to encounter, we consider a strategy close to the Confident Execution proposed in~\cite{Che09}: at each iteration, the confidence of the user is assessed and a query minimizing Eq.~\eqref{eq:MCLU} is presented to him/her only if the confidence about that pixel exceeds a given threshold $\theta$. If the threshold is not met, the query is skipped and the definition of states is upgraded using this negative example. Practically, we train a second model that learns the user's confidence in labeling. This  model learns to separate situations where labeling is feasible (with current knowledge) from other where the user is not supposed to be able to provide a label ($Y_\theta = [-1;+1]$). Contrarily to~\cite{Che09}, we do not make the difference between states that are unfamiliar (that would correspond to new classes) and merely ambiguous.

To interpret the confidence as a probability and normalize it across iterations of the active learning loop, the outputs of the model are converted into probabilities: this operation is natural for models as LDA or neural networks, but when using SVM (as in this study) an estimation as the one proposed by Platt~\cite{Pla99} has to be used. 

Algorithm~\ref{alg:AL-UC} summarizes the flowchart of the proposed method. Note that in AL-UC there are two training sets: the first is the usual set of the classifier with output space $Y = [1, ..., \Omega]$, while the second is a training set containing the confidence samples and a binary output $Y_\theta = [-1;1]$. If at the beginning the input samples coincide ($\x^\epsilon = \x^\epsilon_\theta, \epsilon = 1$), they start to diverge as soon as bad states are encountered: in this case, the batch of training examples for the multiclass classifier is not updated, but the uncertain sample is added to the confidence classifier as a negative example (line~\ref{aa} of the Algorithm) . This way the confidence classifier is constantly updated as long as bad states are encountered. For this reason $|X^\epsilon| \leq |X^\epsilon_\theta|$.

\begin{algorithm}[!t]
\caption{AL-UC algorithm}
\label{alg:AL-UC}
\vspace{0.2cm}
\textbf{Inputs}\\
- Initial training set $X^\epsilon = \{\x_i,y_i\}_{i=1}^l$\\
\white{-}  ($X \in \mathbb{R}^d$, $Y \in [1, ..., \Omega]$, $\epsilon=1$, $\epsilon=$ iteration).\\
- Initial confidence training set $X^\epsilon_{\theta} = \{\x_i,\boldsymbol{1}_i\}_{i=1}^l$\\
\white{-}  ($X \in \mathbb{R}^d$, $\epsilon=1$).\\
- Pool of candidates $U^\epsilon = \{\x_i\}_{i=l+1}^{l+u}$ ($U \in \mathbb{R}^d$, $\epsilon=1$).\\
- Number of pixels $m$ to add at each iteration\\
\white{-} (defining the size of the batch of selected pixels $S$).
\vspace{0.2cm}
\begin{algorithmic}[1]
\REPEAT 
	\STATE Train the classifier with current training set $X^\epsilon$;
	\STATE Train the confidence classifier with current $X^\epsilon_\theta$;
	
	\FOR{each candidate in $U^\epsilon$}
		\STATE Evaluate the active learning \emph{heuristic};
		\STATE Assess assignment's confidence $p(y_\theta = +1| X^\epsilon_\theta)$;
	\ENDFOR
	
	\STATE Rank the candidates in $U^\epsilon$ according to the score of the heuristic, obtain ranking $r$;

	\REPEAT
		\STATE Select next candidate in $r$, $\x_r$; 
		\IF{$p(y_{\theta,r} = +1 | X^\epsilon_\theta) > \theta$}
		\IF {the user can provide the label $y_r$}
			\STATE Add the labeled candidate to the batch $S^\epsilon = S\epsilon \cup \{\x_r, y_r\}$;
			\STATE Add the positive example to the confidence training set $X^\epsilon_\theta = X^\epsilon_\theta \cup \{\x_r, 1\} $;
			\ELSE 
				\STATE Add the negative example to the confidence training set $X^\epsilon_\theta = X^\epsilon_\theta \cup \{\x_r, -1\} \label{aa}$;
				\ENDIF

		\ELSE
			\STATE Add the negative example to the confidence training set $X^\epsilon_\theta = X^\epsilon_\theta \cup \{\x_r, -1\} \label{ab}$;
		\ENDIF
	\UNTIL{Batch $S$ has $m$ candidates}
	
	 \STATE Add the batch to the training set $X^{\epsilon+1}=X^{\epsilon} \cup S^\epsilon$;
	\STATE Remove the batch from the pool of candidates $U^{\epsilon+1}=U^{\epsilon} \backslash S^\epsilon$;
	\STATE $\epsilon = \epsilon + 1$.
			
\UNTIL{a stopping criterion is met.}
\end{algorithmic}
\vspace{0.2cm}
\end{algorithm}

\section{Data and setup}~\label{sec:data}
In this section, we present the datasets considered in the experiments, as well as the general experimental setup adopted.

\subsection{Datasets}
\begin{figure}[t]
\centering
\begin{tabular}{cc}
\includegraphics[width = 4cm]{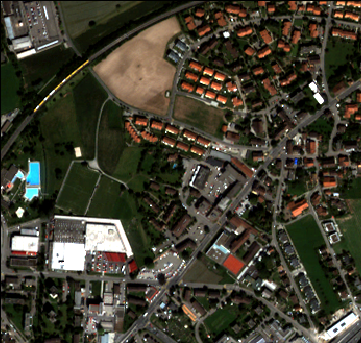}&
\includegraphics[width = 4cm]{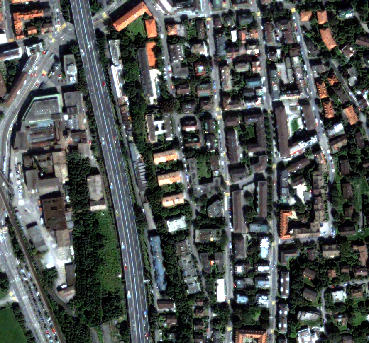}\\
\includegraphics[width = 4cm]{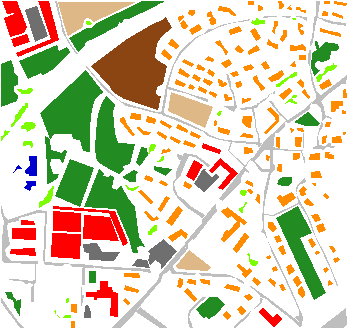}&
\includegraphics[width = 4cm]{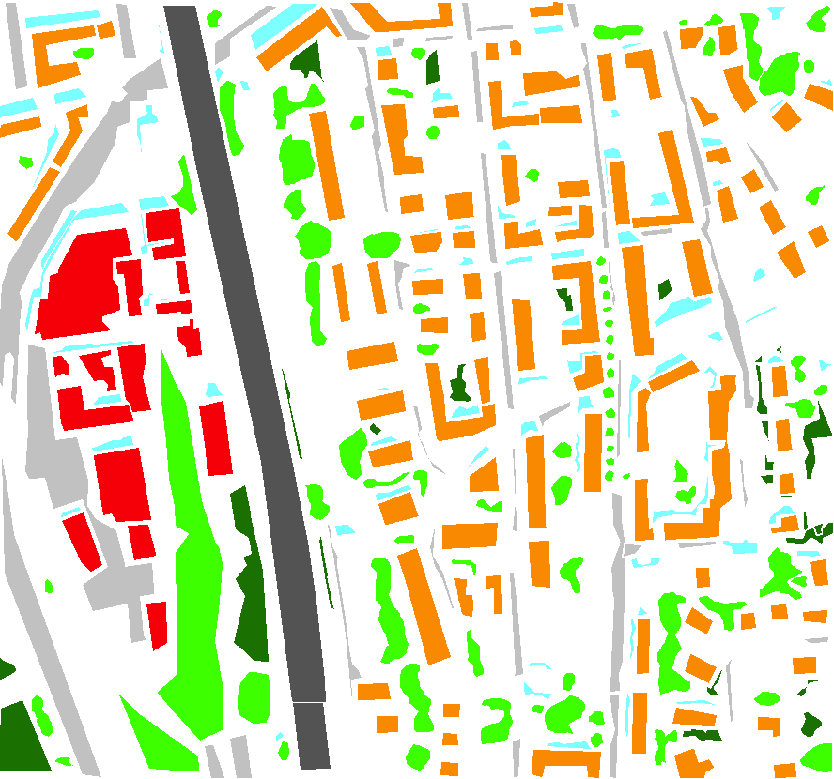}\\
(a) Br\"uttisellen & (b) Highway\\
\end{tabular}
\caption{Images considered in the experiments, along with their corresponding GTs used for testing purposes.}
\label{fig:imgs}
\end{figure}

Two urban VHR images are considered for the experiments (Fig.~\ref{fig:imgs}). They describe urban environments at different spatial resolutions, in order to assess differences in confidence of labeling related to the resolution of objects.
\begin{itemize}
\item[-] QuickBird ``Br\"uttisellen''. The first image is a 4-bands optical image of a residential neighborhood of the city of Zurich named Br\"uttisellen acquired by the sensor QuickBird in 2002. The image has a size of $329\times 347$ pixels, and a geometrical resolution of 2.4~m. Nine classes of interest have been highlighted by photointerpretation and $40,762$ pixels are available (see Tab~\ref{tab:GT}).
\item[-] QuickBird ``Highway''. The second image is another 4-bands optical image of an industrial neighborhood of the city of Zurich acquired by the sensor QuickBird in 2006. The image has a size of $828\times 889$ pixels. The original image was pansharpened using Bayesian Data Fusion~\cite{Fas08} to attain a spatial resolution of 0.6~m. Seven classes of interest have been highlighted by photointerpretation and $254,469$ pixels are available (see Tab~\ref{tab:GT}).
\end{itemize}

To account for the spatial context of the pixel, we stacked morphological features~\cite{Soi04} to the spectral vector: we added opening and closing features computed on the first PCA extracted on the multispectral image, as in~\cite{Ben05,Lic09}\blue{, which is a valid alternative to the use of a panchromatic image (as in~\cite{Lon12,Mur10})}. This operation allows to separate land use classes of similar materials but with different spatial extents as, for instance, roads and parking lots. Since the images have different spatial resolutions, the structuring element sizes are $ \{1,3\}$ in radius for the Br\"uttisellen image and $\{3,6,9\}$ for the Highway image. Shape is kept circular in both cases.

The pixels highlighted by photointerpretation compose the test set on which the different approaches are evaluated. Their specific quantities are highlighted in Tab.~\ref{tab:GT}.

\begin{table}[!t] \centering
\caption{Number of labeled pixels used for evaluation of the Br\"uttisellen and Highway images.}
\label{tab:GT}
\begin{tabular}{c|c|c|p{1.6cm}}
\hline
Image&Class & GT pixels & Legend color\\
\hline\hline
\multirow{8}{*}{Br\"uttisellen}
& Trees                 & $1,095$  & Light green \\
& Meadows               & $13,123$ & Dark green \\
& Harvested vegetation  & $2,523$  & Light brown \\
& Bare soil             & $3,822$  & Brown \\
& Residential buildings & $6,746$  & Orange \\
& Commercial buildings  & $5,277$  & Red \\
& Asphalt               & $6,158$  & Light gray \\
& Parkings              & $1,749$  & Dark gray \\
& Pools                 & $269$    & Blue \\

\hline\hline
\multirow{7}{*}{Highway}
& Trees                 & $52,813$ & Light green \\
& Meadows               & $12,347$ & Dark green \\
& Residential buildings & $78,018$ & Orange \\
& Commercial buildings  & $25,389$ & Red \\
& Highway               & $28,827$ & Dark gray \\
& Asphalt               & $43,005$ & Light gray \\
& Shadows               & $14,071$ & Cyan \\
\hline
\end{tabular}
\end{table}

\subsection{Experimental setup}

\begin{figure}[t]
\centering
\begin{tabular}{c}
\includegraphics[width = 8cm]{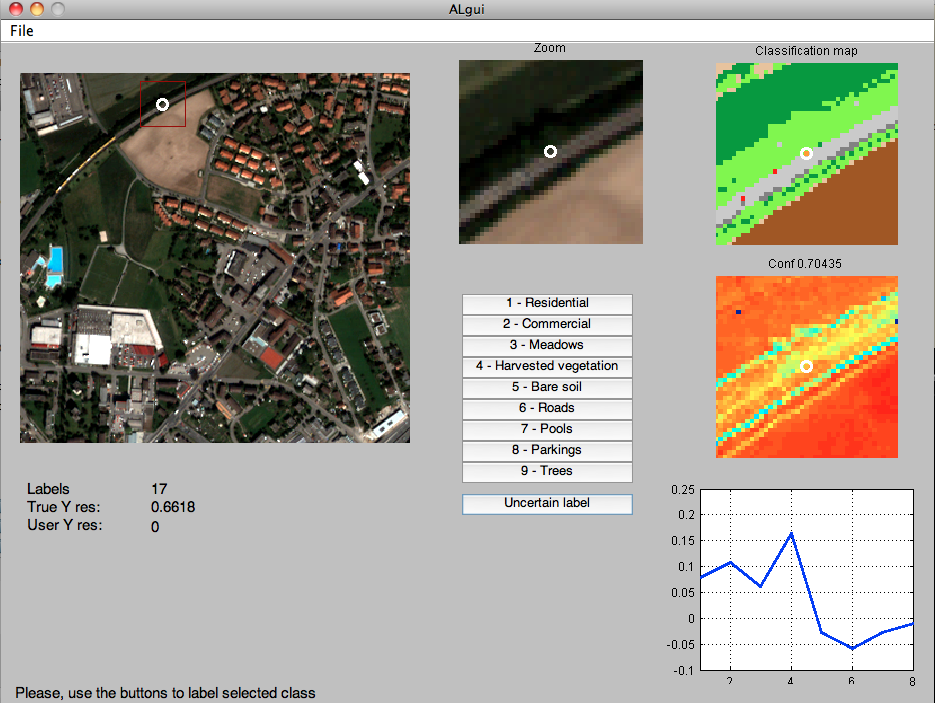}\\
\end{tabular}
\caption{Graphic user interface developed for testing the AL-UC model. On the left, the image to be labeled; in the middle, class buttons and a zoom; on the right the current classification map, the confidence map and the spectrum of the pixel in the white circle.}
\label{fig:gui}
\end{figure}

To test the proposed active learning with user's confidence model we built a MATLAB graphic user interface (Fig.~\ref{fig:gui}), where real users have the task to label the VHR images of Zurich, Switzerland presented above. Contrarily to  active learning papers in remote sensing previously published, all the image can be sampled and a human user is performing the labeling.

\begin{figure*}[!t]
\centering
\begin{tabular}{ccccc}
\rotatebox{90}{Iteration $\epsilon =1$} &
\rotatebox{90}{\hspace{0.2cm } ($|X^1| = 65, |X^1_\theta| = 82$)} &
  \includegraphics[width=4.55cm]{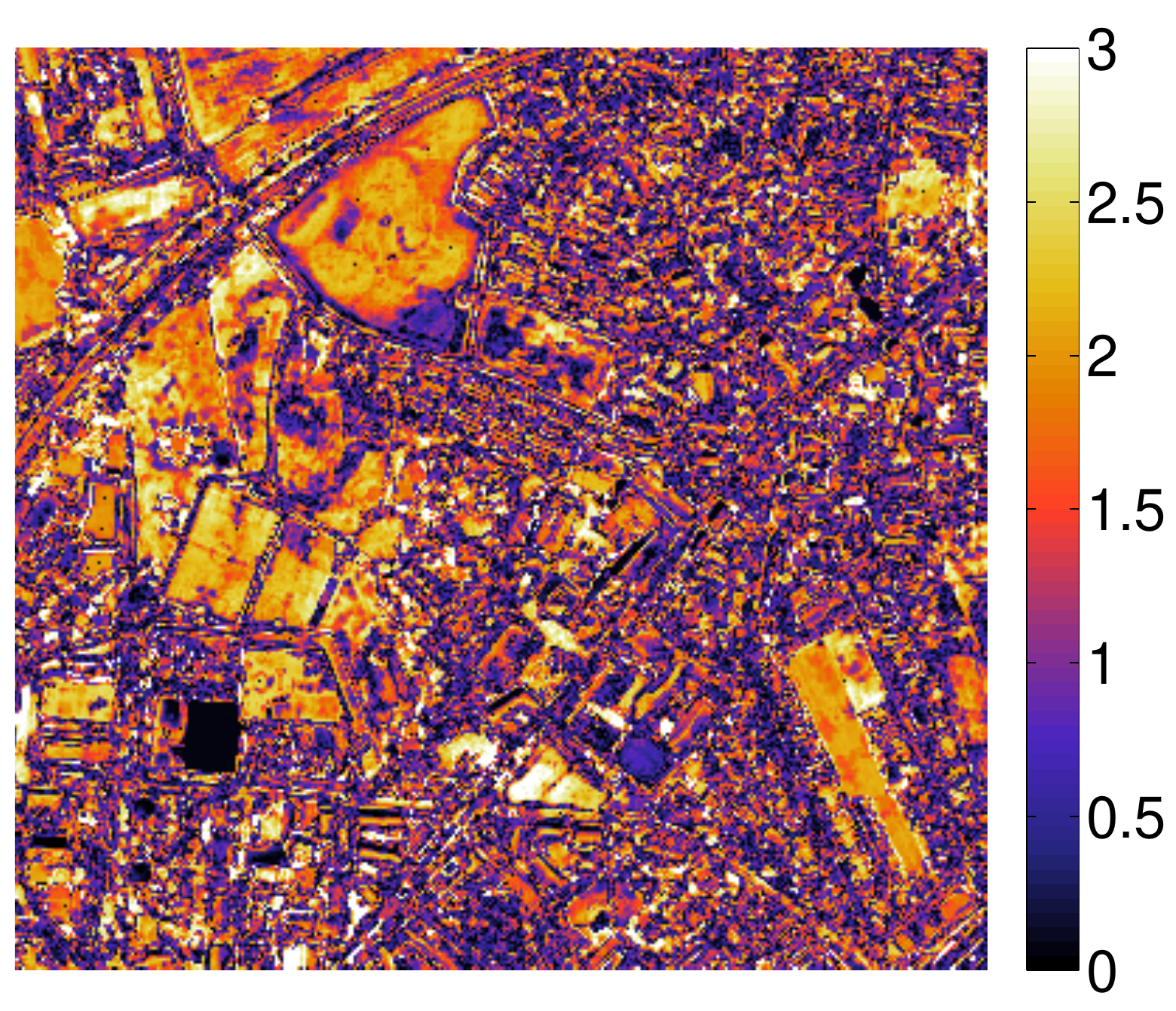} &
  \includegraphics[width=4.55cm]{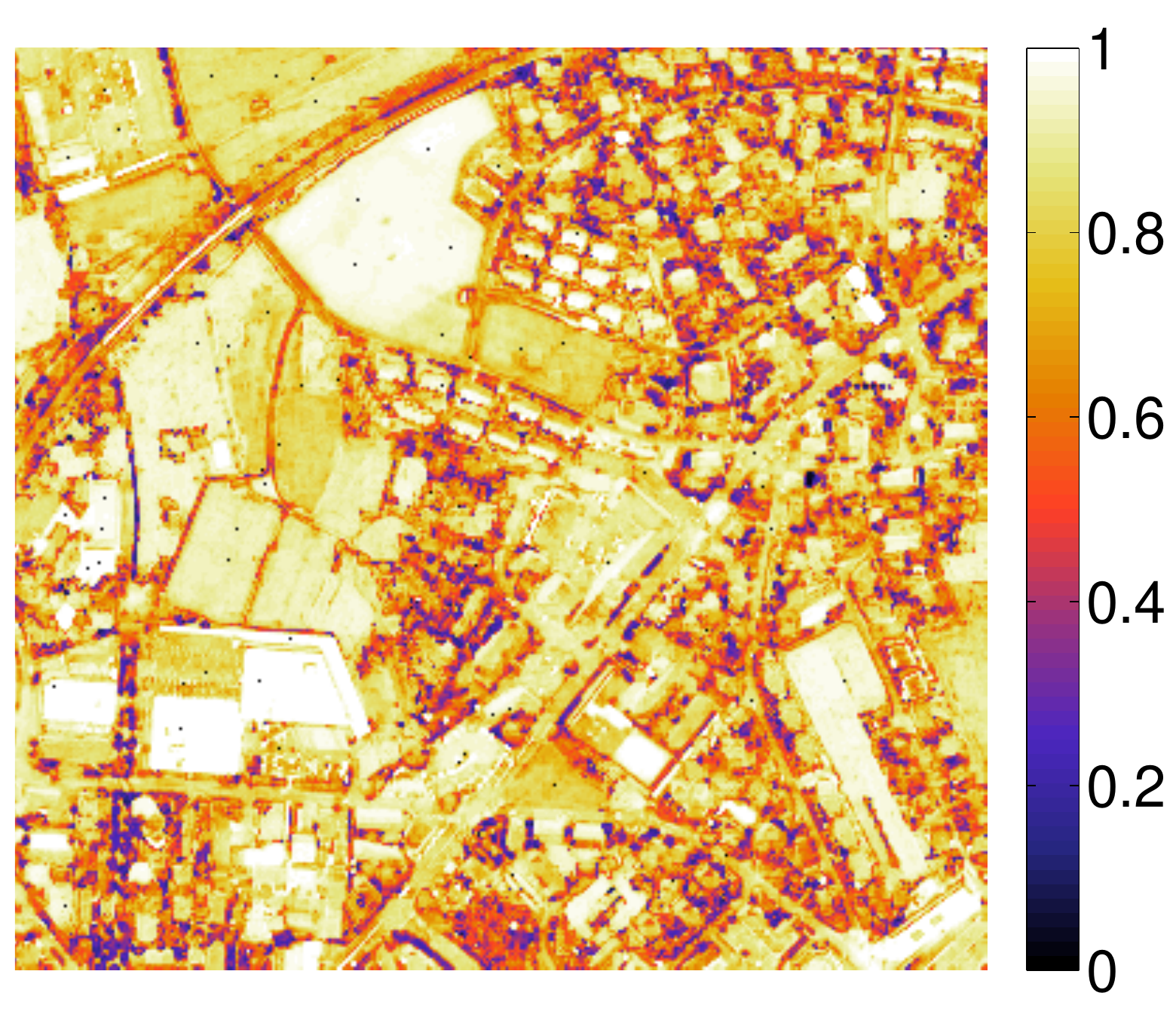} &
  \includegraphics[width=4cm]{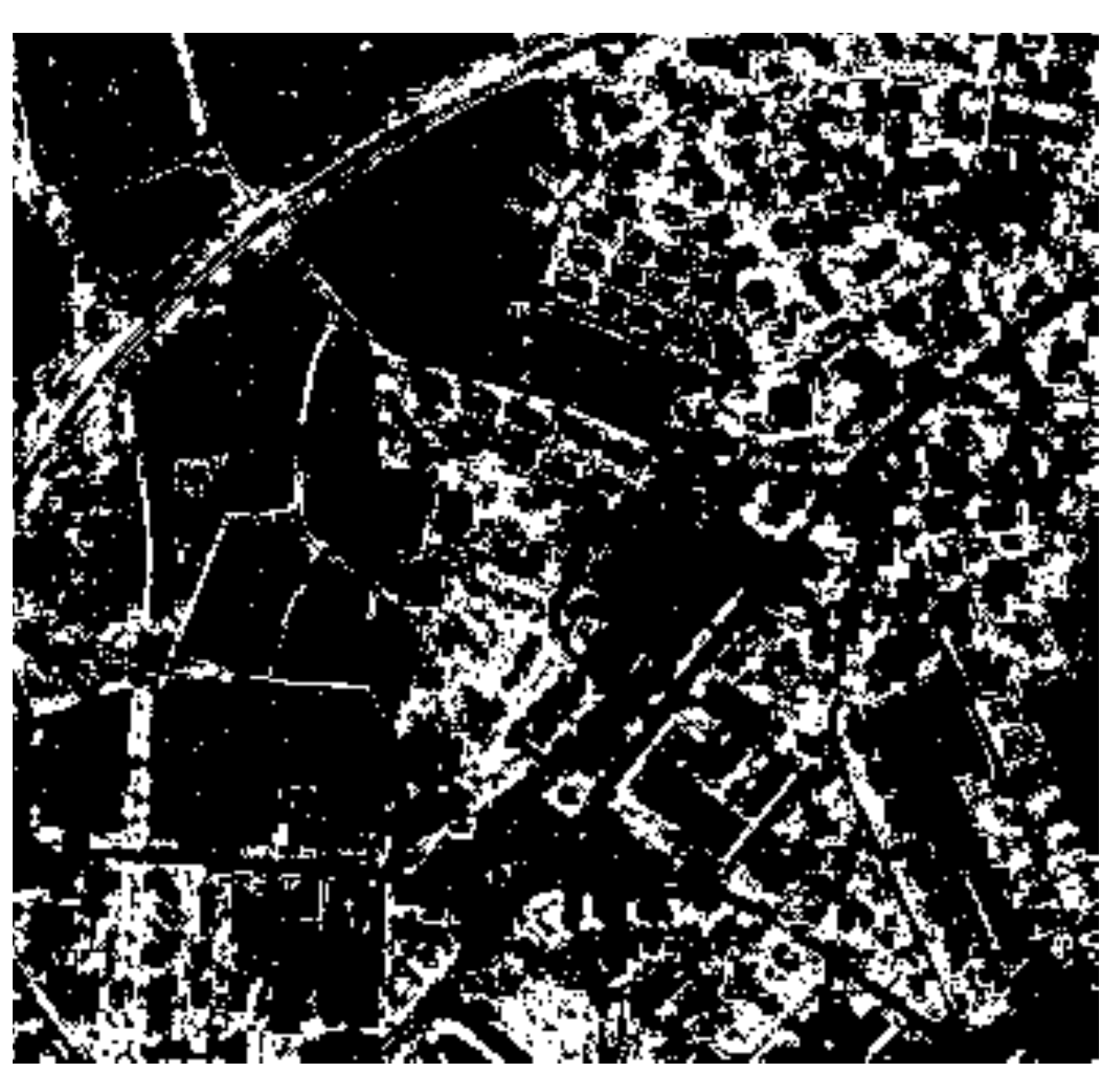} \\
\rotatebox{90}{Iteration $\epsilon = 2$} &
\rotatebox{90}{\hspace{0.2cm } ($|X^2| = 85, |X^2_\theta| = 111$)} &
  \includegraphics[width=4.55cm]{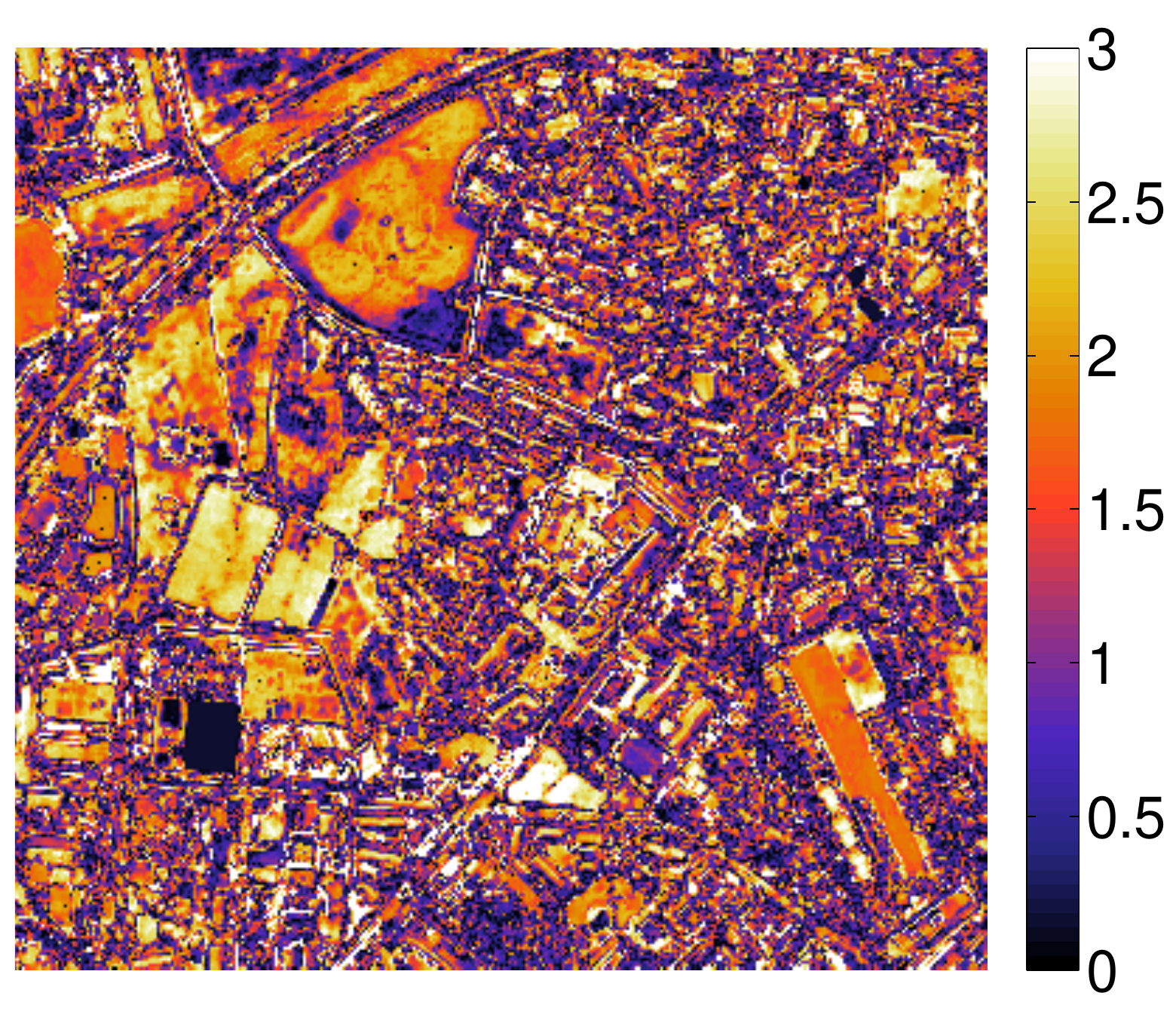} &
  \includegraphics[width=4.55cm]{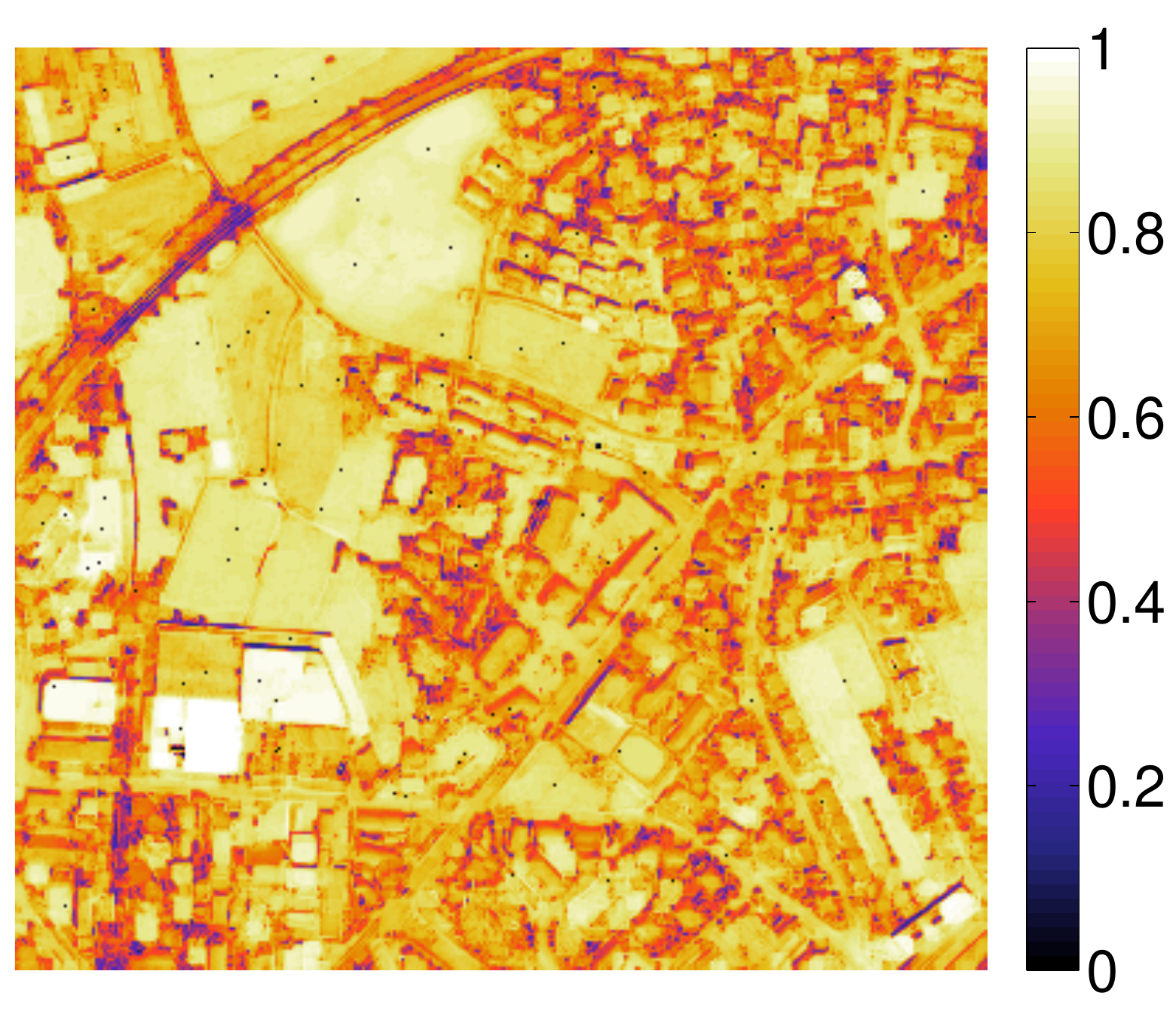} &
  \includegraphics[width=4cm]{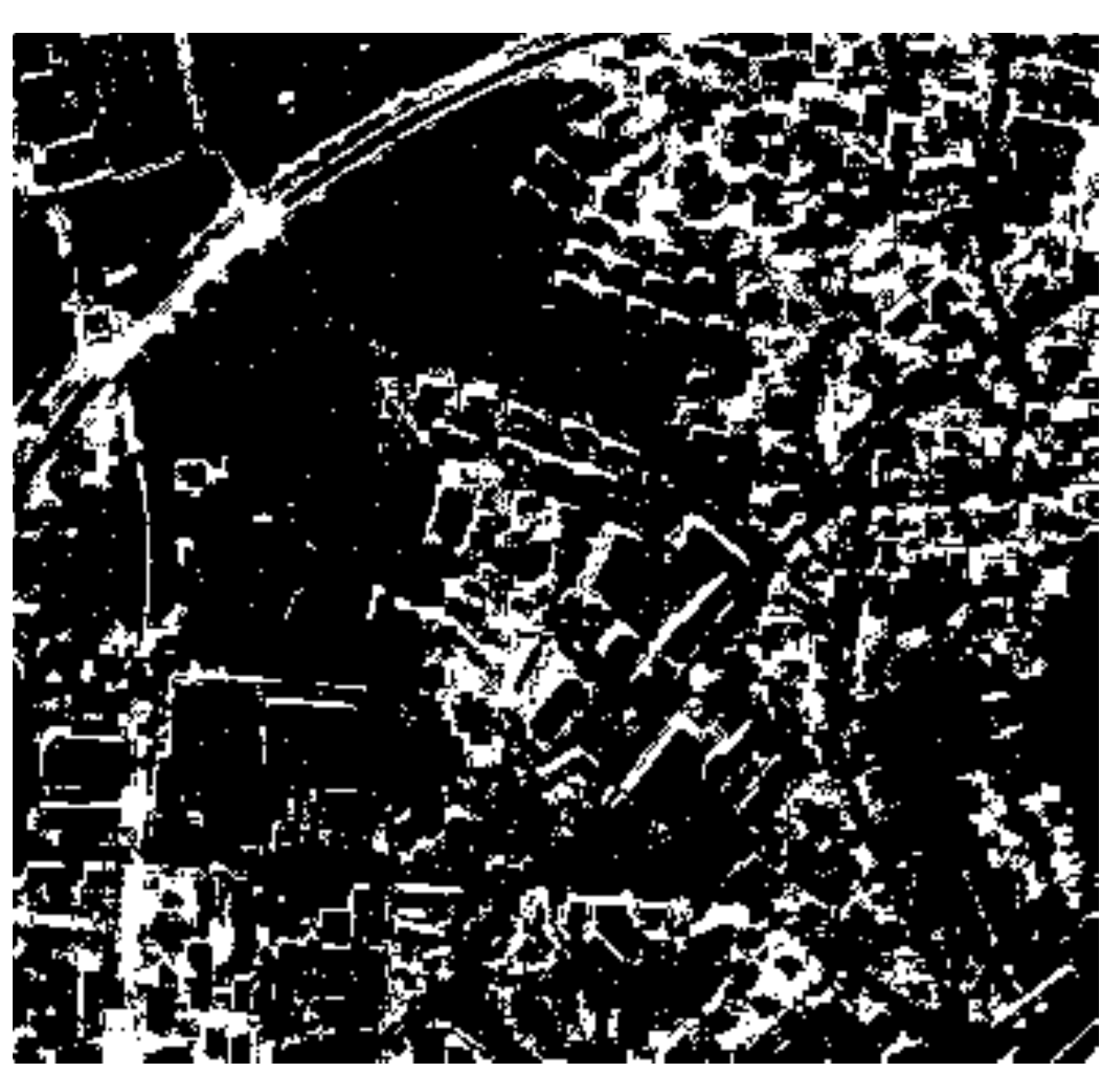} \\
\rotatebox{90}{ Iteration $\epsilon = 5$ } &
\rotatebox{90}{\hspace{0.2cm }($|X^5| = 165, |X^5_\theta| = 207$)} &
  \includegraphics[width=4.55cm]{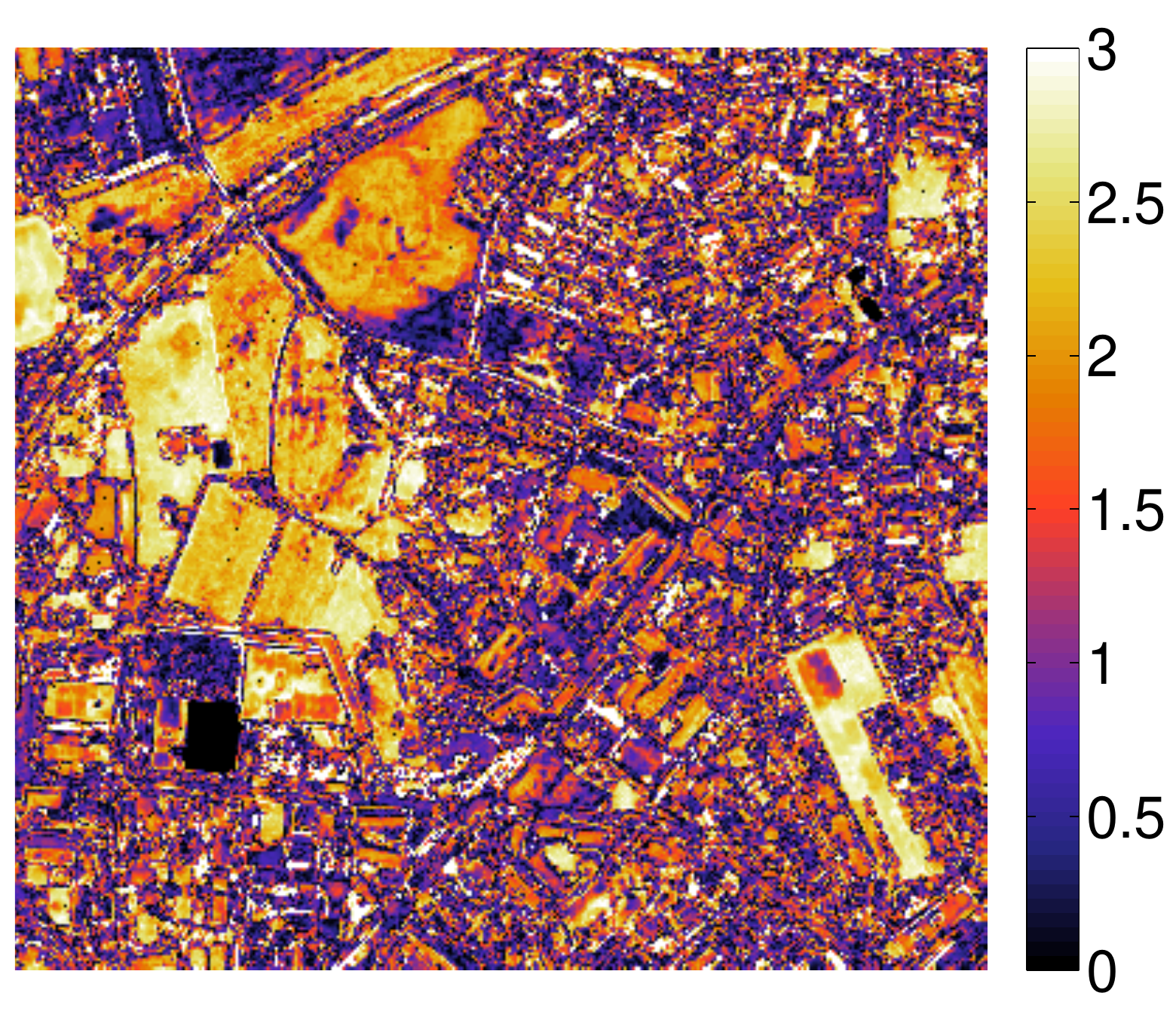} &
  \includegraphics[width=4.55cm]{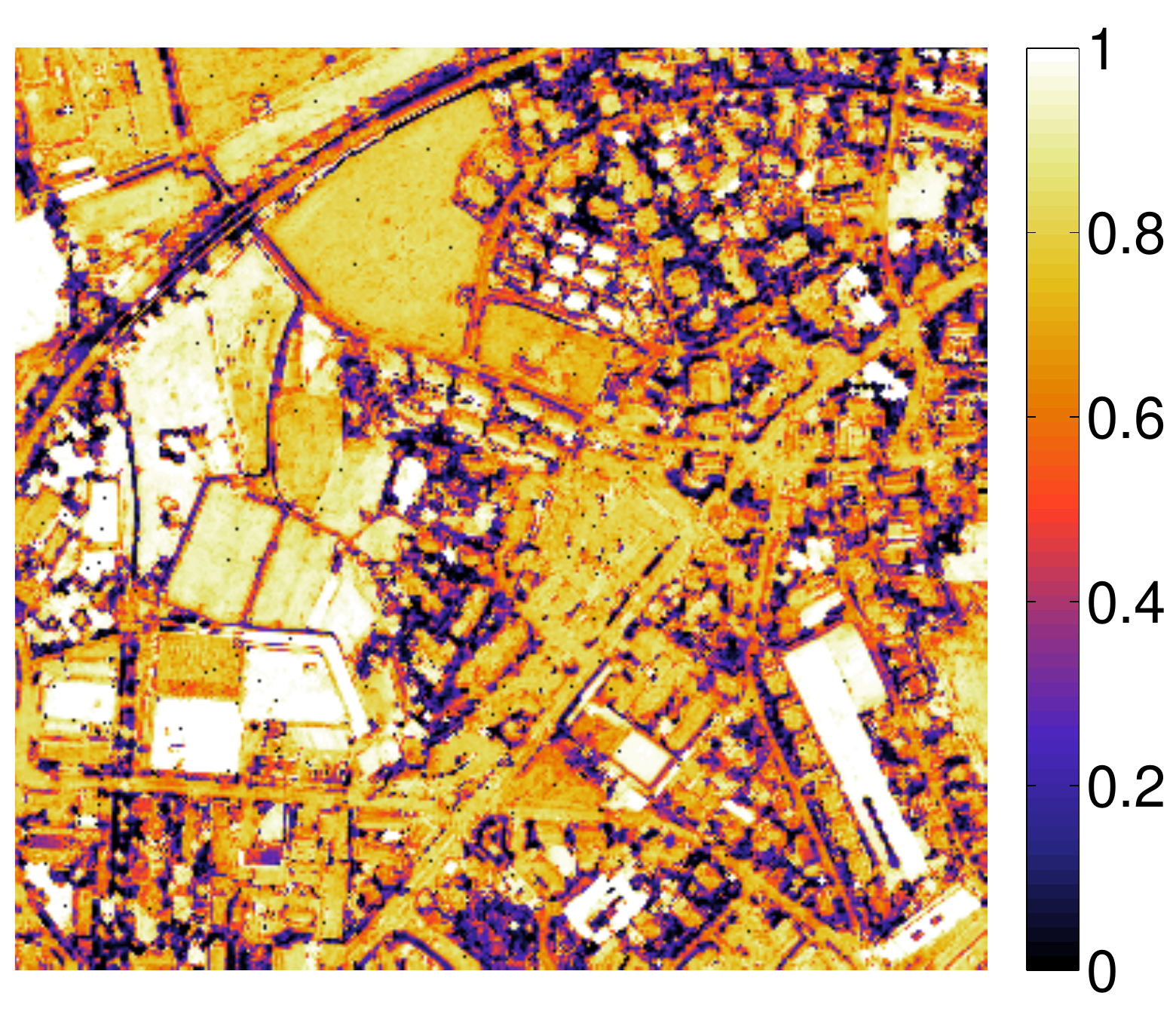} &
  \includegraphics[width=4cm]{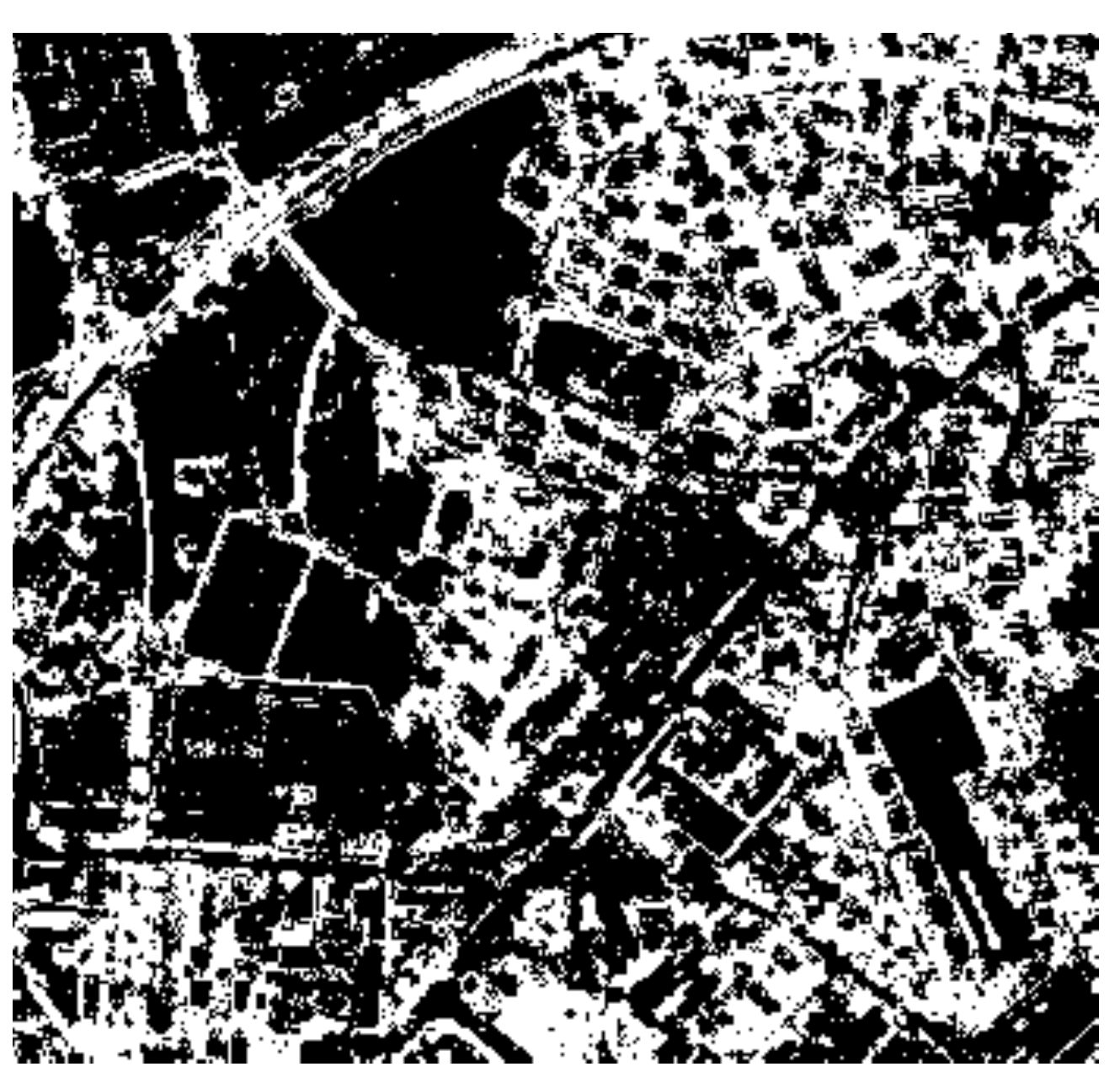} \\
& &(a)  AL criterion & (b) Confidence map & (c) Confidence mask, $\theta = 0.6$ \\
\end{tabular}
\caption{\blue{AL-UC ingredients. (a) A classical active learning heuristic: MCLU. [Uncertain samples correspond to dark colors] (b) A user's confidence estimation: SVM trained on user's responses [Easily labelizable samples correspond to light colors]. (c) Mask obtained by thresholding the confidence map at $\theta = 0.6$. Only pixels in the black areas can now be selected by minimizing the MCLU criterion.}}
\label{fig:maps}
\end{figure*}

After initialization, the user is invited to enter an initial training set by photointerpretation. As for their difference in size, 5 pixels per class are queried for the ``Br\"uttisellen'' image ($|X^1| = 45$), while 10 pixels per class are requested for the ``Highway'' image ($|X^1| = 70$). He/she is invited to choose these pixels on the image through an interactive window. These samples receive a label by the user and a positive label for the confidence classifier. Then, the active learning process starts.

The user is asked to label the pixels selected by an active learning algorithm into one of the $\Omega$ classes of interest ($Y = [1, ..., \Omega]$), detailing the different urban land use types, or into an ``unknown class'', if the user does not know which label to assign. Every 20 valid labeled pixels (i.e. every 20 answers other than ``unknown class'') the classifier retrains with the increased training set and produces a new raking of the unlabeled pixels. \blue{For EQB, we considered a committee of $10$ models, each one using $75\%$ of the available training data drawn randomly from $X^\epsilon$.}

As a base classification model, we used a nonlinear SVM with RBF Gaussian kernel. As a lower bound of performance, we used a random selection of locations (RS). Since the first training set $X^1$ is very small, the RBF kernel parameter is estimated using the median distance between pixels in the image. The Torch library is used for the multiclass SVM~\cite{Col02}, which implements the one-against-all (OAA) strategy.

To avoid bad states on the active learning output, we learn the user's confidence by training a second binary SVM classifier with RBF kernel. We consider the probabilistic output using Platt's method~\cite{Pla99}. The LibSVM solver is used, as it returns these posterior probabilities. This classifier is trained only since iteration $\epsilon = 2$, because at the first iteration there are no negative samples ($X^1_\theta$  contains only positive samples chosen by the user). To avoid overfitting of specific situations, we kept the search range for the $\sigma$ kernel bandwidth large in a 4-folds crossvalidation strategy ($\sigma = [10^{-1}, ..., 10^3]$). The threshold $\theta$ is fixed at $0.6$ after experimental testing: it constitutes a good tradeoff between a filtering that is too strong (which would return uninformative pixels) and one that is too weak (which would be identical to common AL).

\blue{In the experiments reported in Sections~\ref{sec:Bru} and~\ref{sec:High}, a single user  performed $5$ independent experiments, where he choose the initial training  set by clicking on the image.  The user knows the image, as well as the task to be performed (i.e. the ground truth). On the contrary, in Section~\ref{sec:users} we considered different users in the labeling task. In this case, we compared the performance of five users, three with experience in remote sensing and labeling tasks, and two who are not familiar with those tasks. Each user performed a single experiment with the three models (random, standard active learning and the proposed AL-UC, both with MCLU as a base heuristic).} 
As stated above, the different models are compared in terms of estimated Kappa statistics on the entirety of labeled pixels shown in bottom row of Fig.~\ref{fig:imgs} (quantities in Table~\ref{tab:GT}).

\section{Results}\label{sec:res}
This section reports the experimental results obtained on the two case studies.

\subsection{Zurich Br\"uttisellen}\label{sec:Bru}
Figure~\ref{fig:maps} illustrates the basic components of the proposed AL-UC on the Br\"uttisellen dataset: in the left column, the AL heuristic is reported at iterations $1$, $2$ and $5$. Even if during the iterations there is an increase in confidence on large spatially smooth areas, the heuristic remains fragmented in complex areas and many minima can be seen. Central column of Fig.~\ref{fig:maps} illustrates the posterior probability of the confidence classifier: during the iterations, this classifier specializes in detecting areas that are easily recognized by the user and not only large, smooth areas (for example, note how the roads become more and more confident, even if they are thin elongated structures). 

\begin{figure}[!b]
\centering
\begin{tabular}{cc}
\includegraphics[height = 3.5cm]{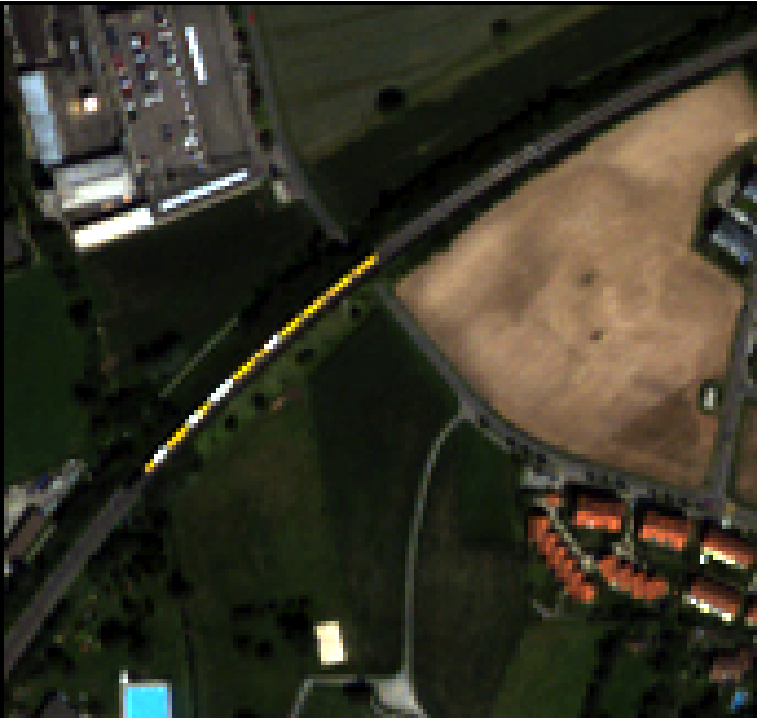}&
\includegraphics[height = 3.8cm]{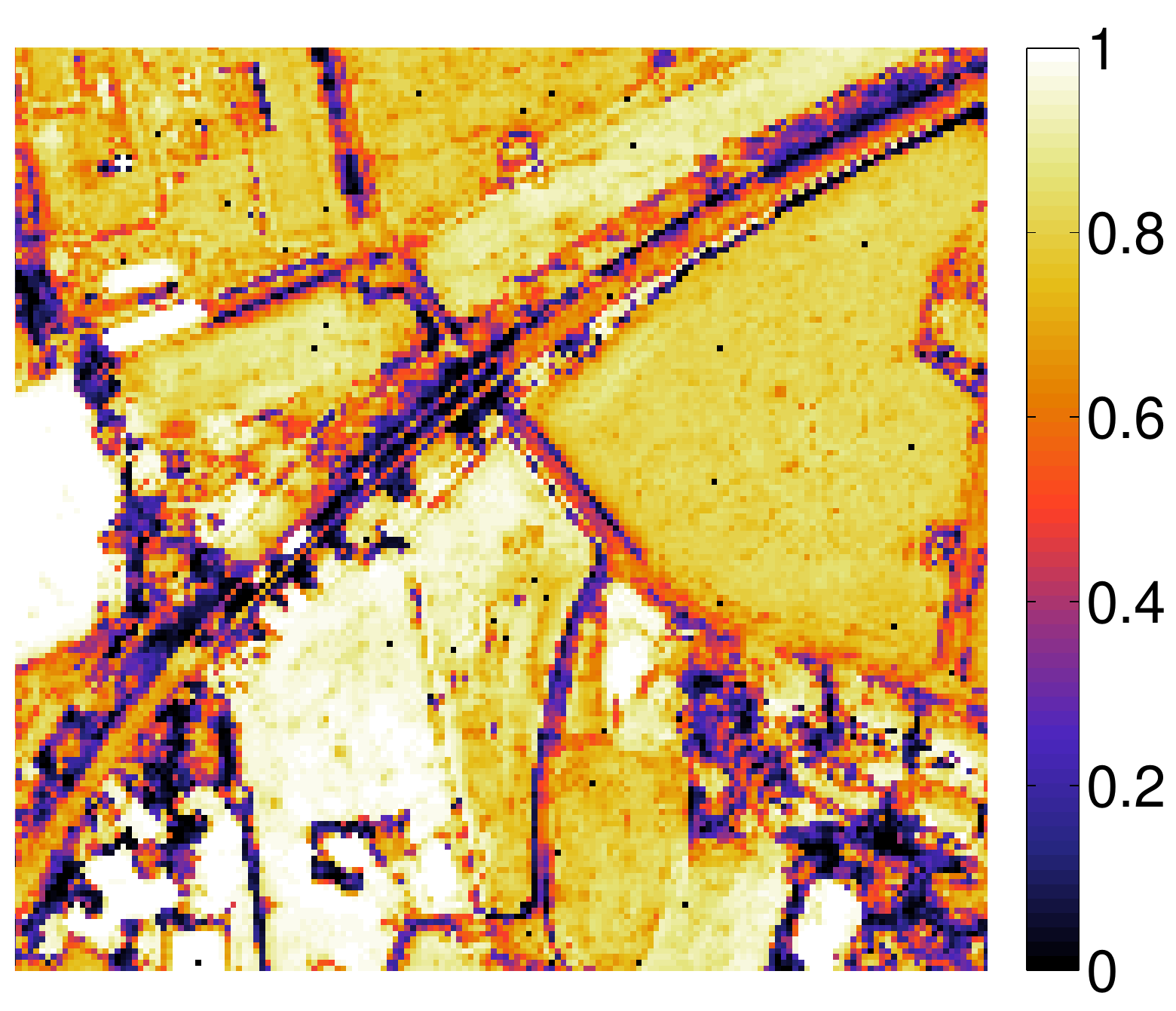}\\
\includegraphics[height = 3.5cm]{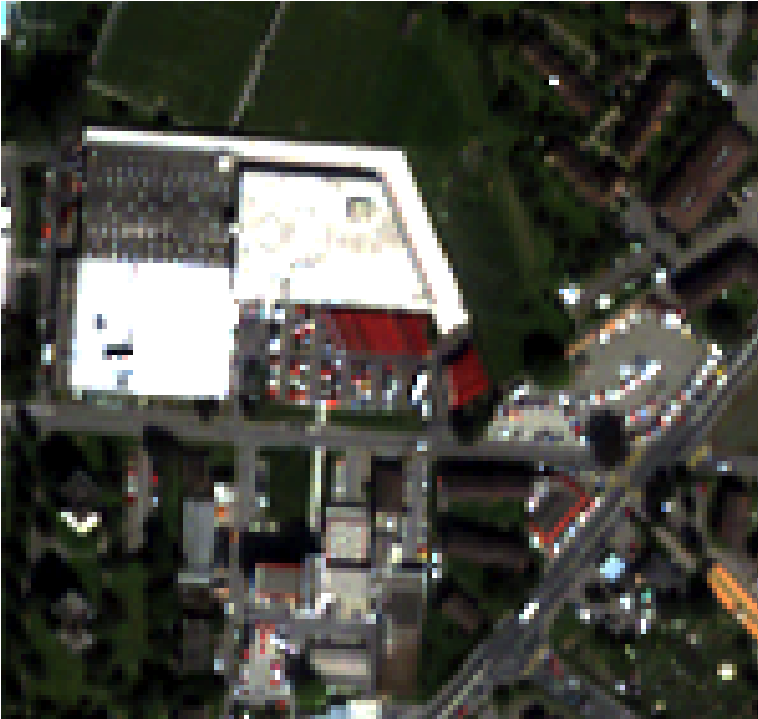}&
\includegraphics[height = 3.8cm]{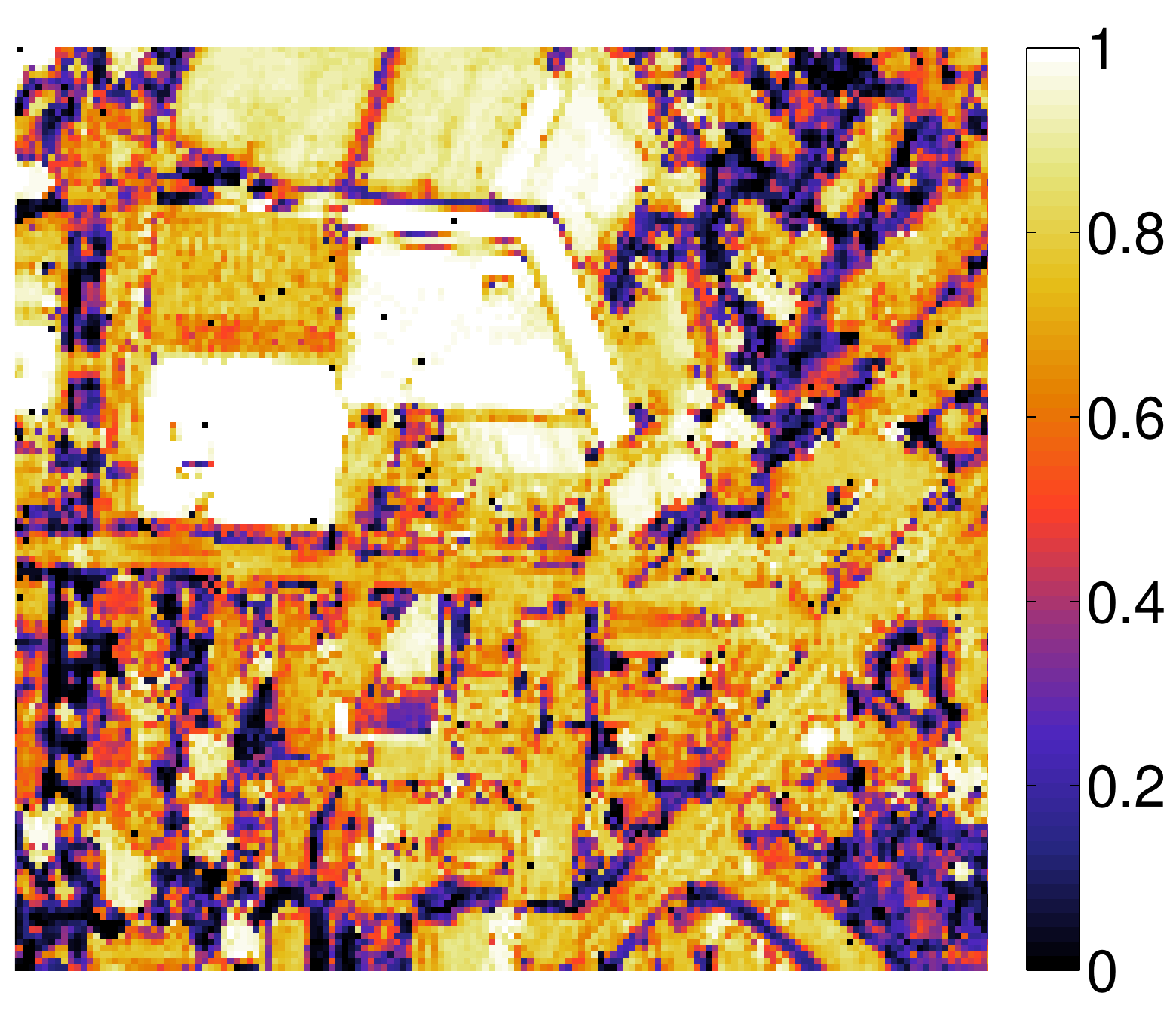}\\
(a) Image & (b) Confidence map\\
\end{tabular}
\caption{Zooms into the confidence map of Fig.~\ref{fig:maps}b at iteration $\epsilon = 5$.}
\label{fig:zooms}
\end{figure}

In Fig.~\ref{fig:zooms}, we show two examples of how the confidence map works: in the top example, the railway is easily labeled by the user and result in high confidence, while the train (in white/yellow) is not among the classes to be detected, so it is handled as a bad state with low confidence. In the bottom example, one can appreciate how the linear structures such as roads receive high confidence, as well as the parking lots, which are characterized by high variance (parked cars). 

\begin{figure*}[!t]
\centering
\begin{tabular}{cccc}
\rotatebox{90}{\hspace{1cm}MCLU~\cite{Dem10}}&
\includegraphics[width=5.3cm]{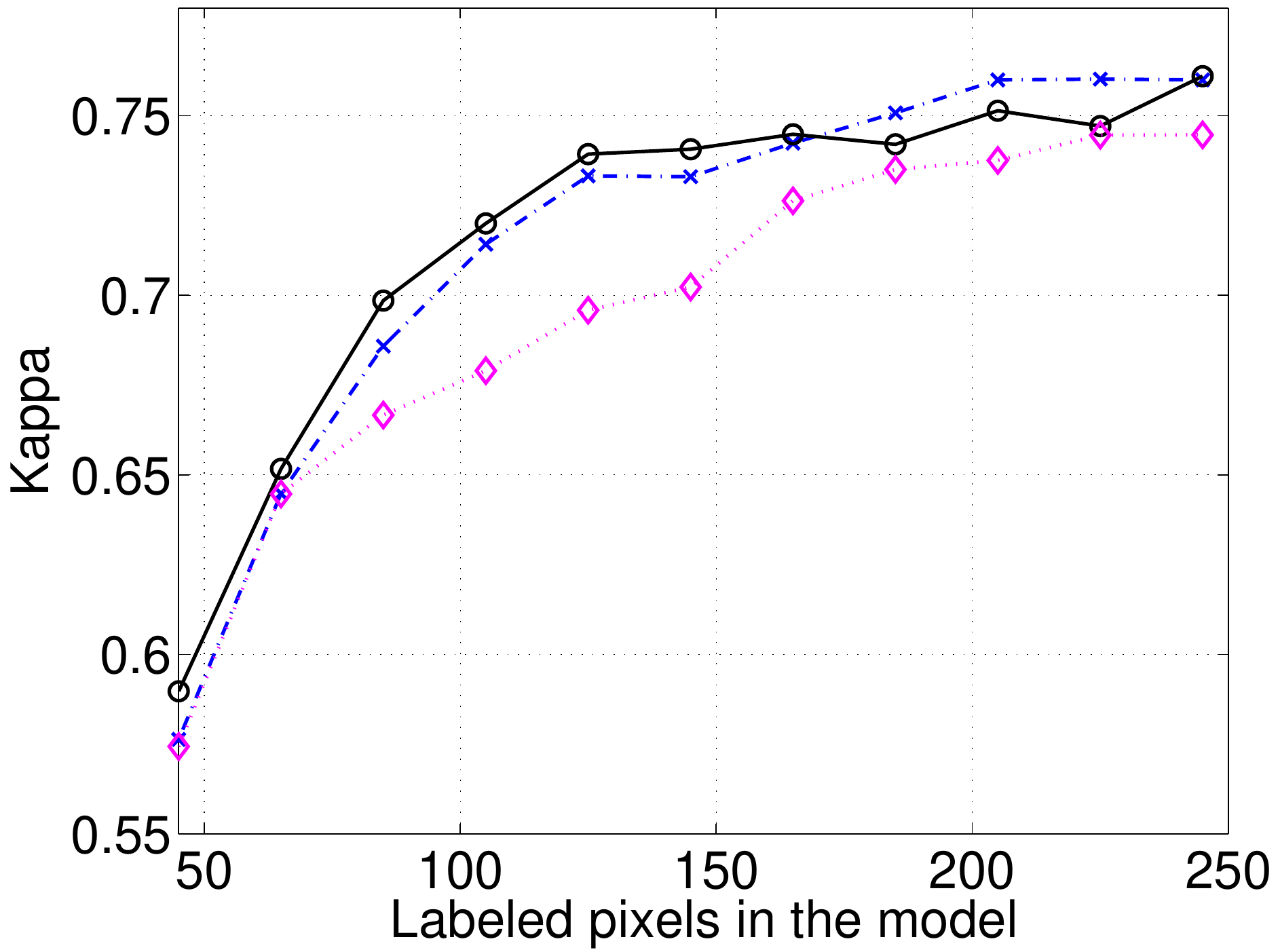}&
\includegraphics[width=5.3cm]{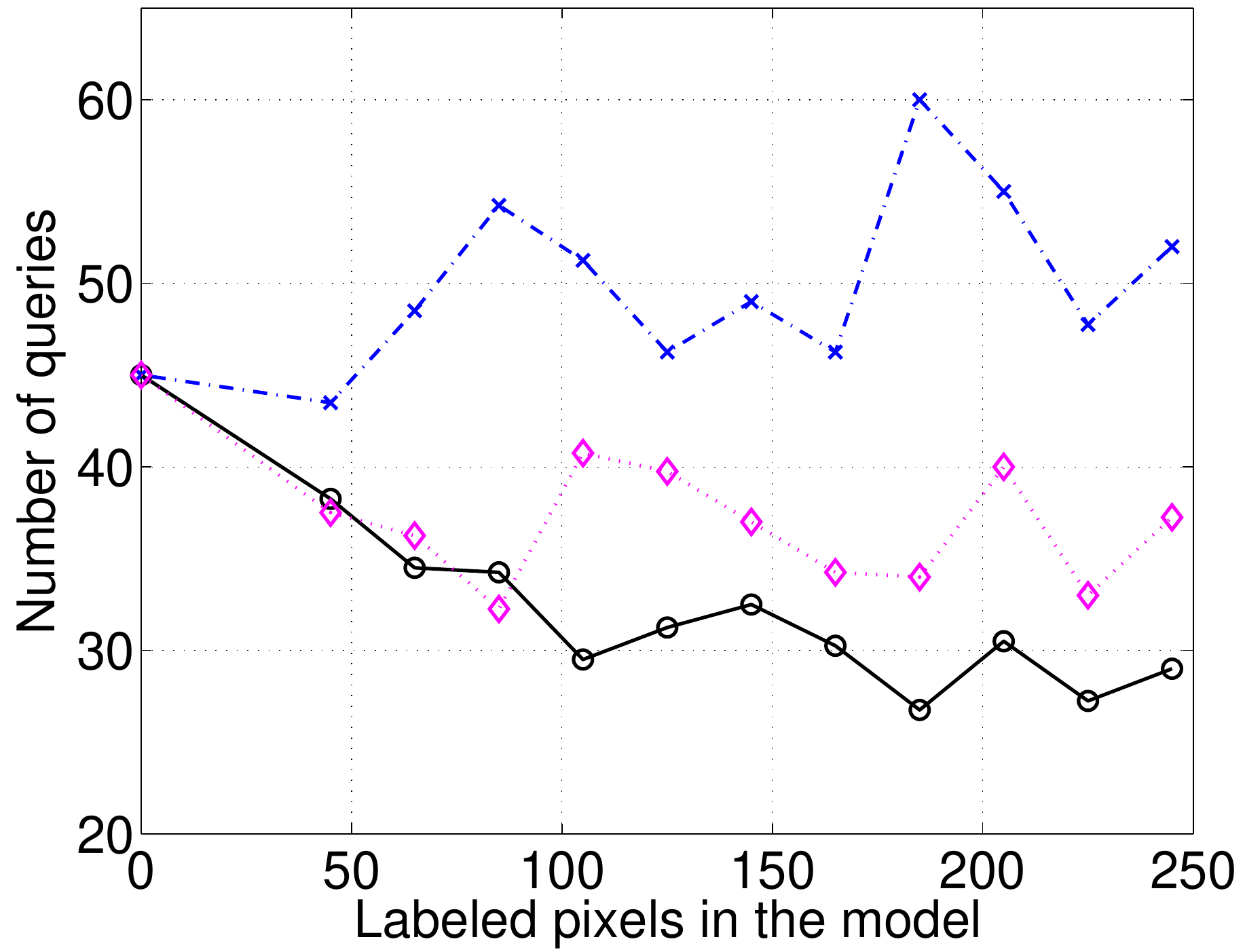}&
\includegraphics[width=5.3cm]{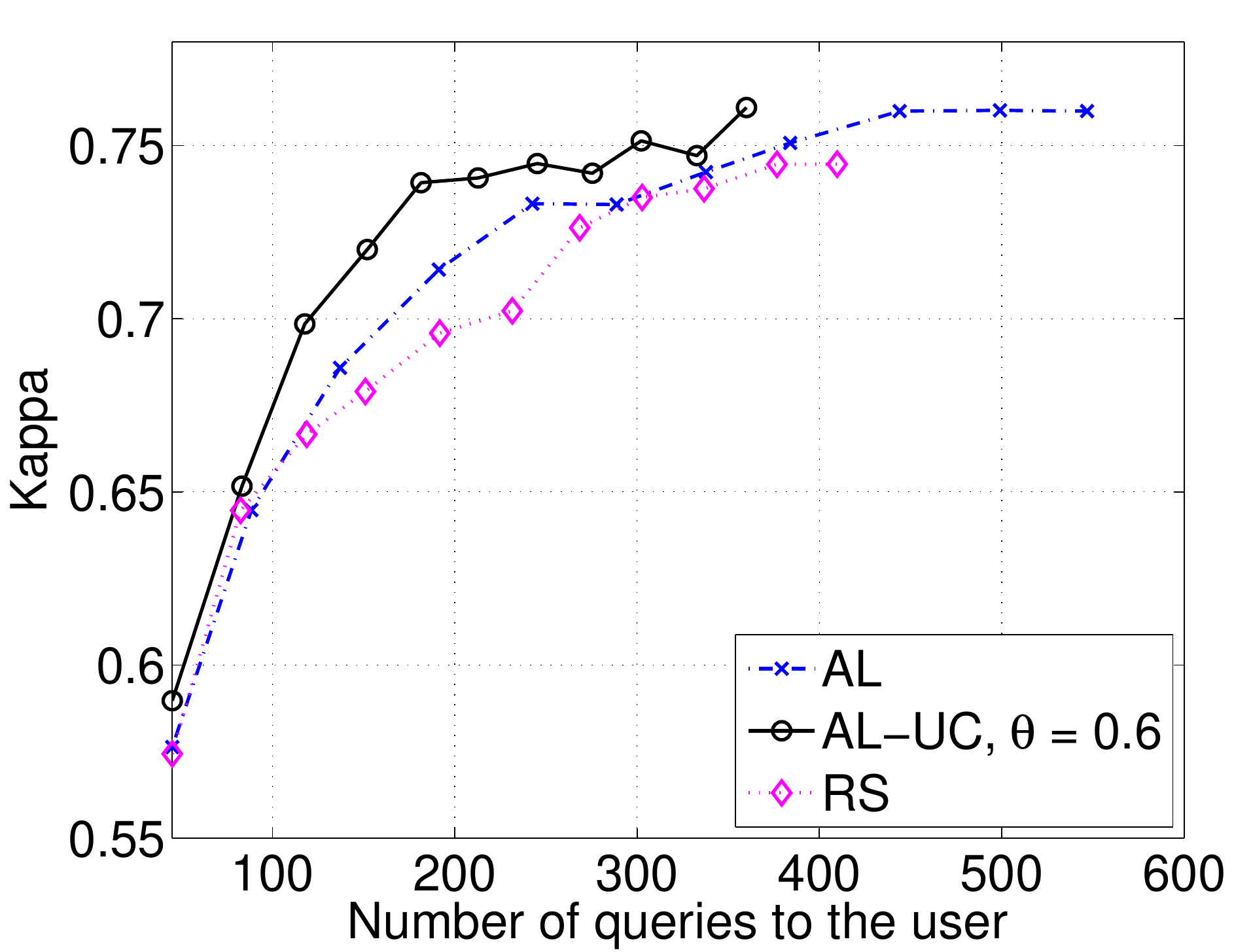}\\
\rotatebox{90}{\hspace{1.8cm}\blue{EQB~\cite{Tui09}}}&
\includegraphics[width=5.3cm]{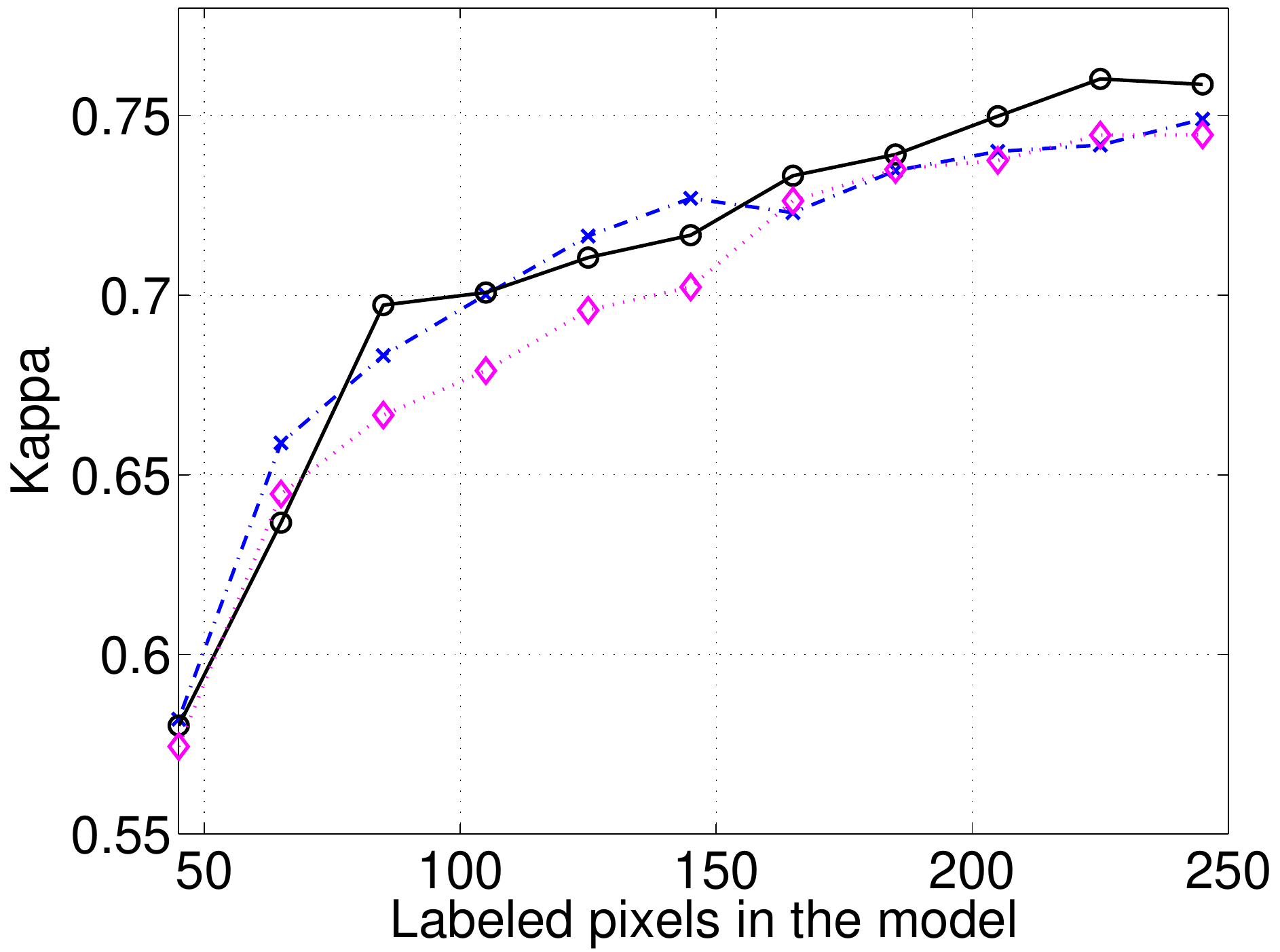}&
\includegraphics[width=5.3cm]{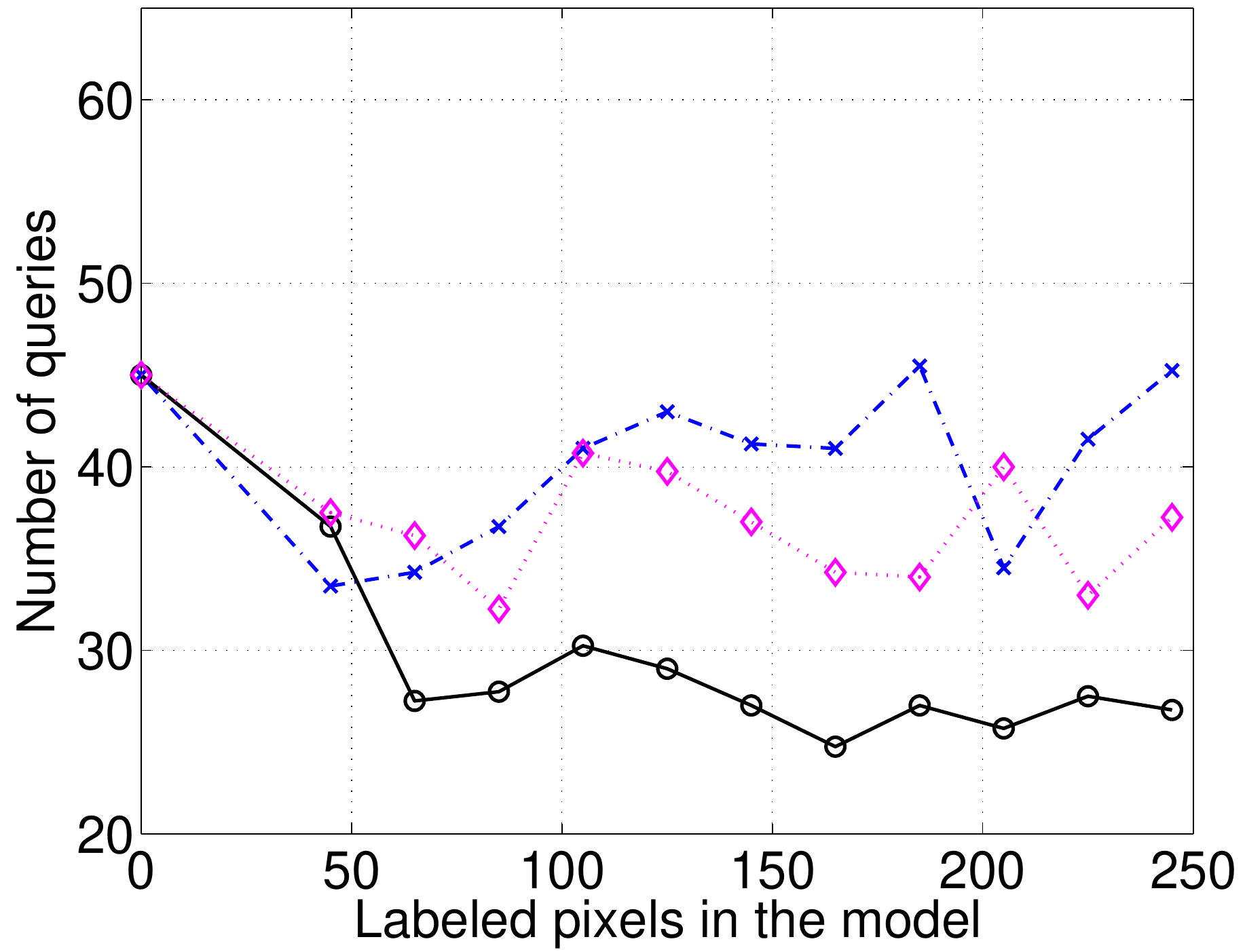}&
\includegraphics[width=5.3cm]{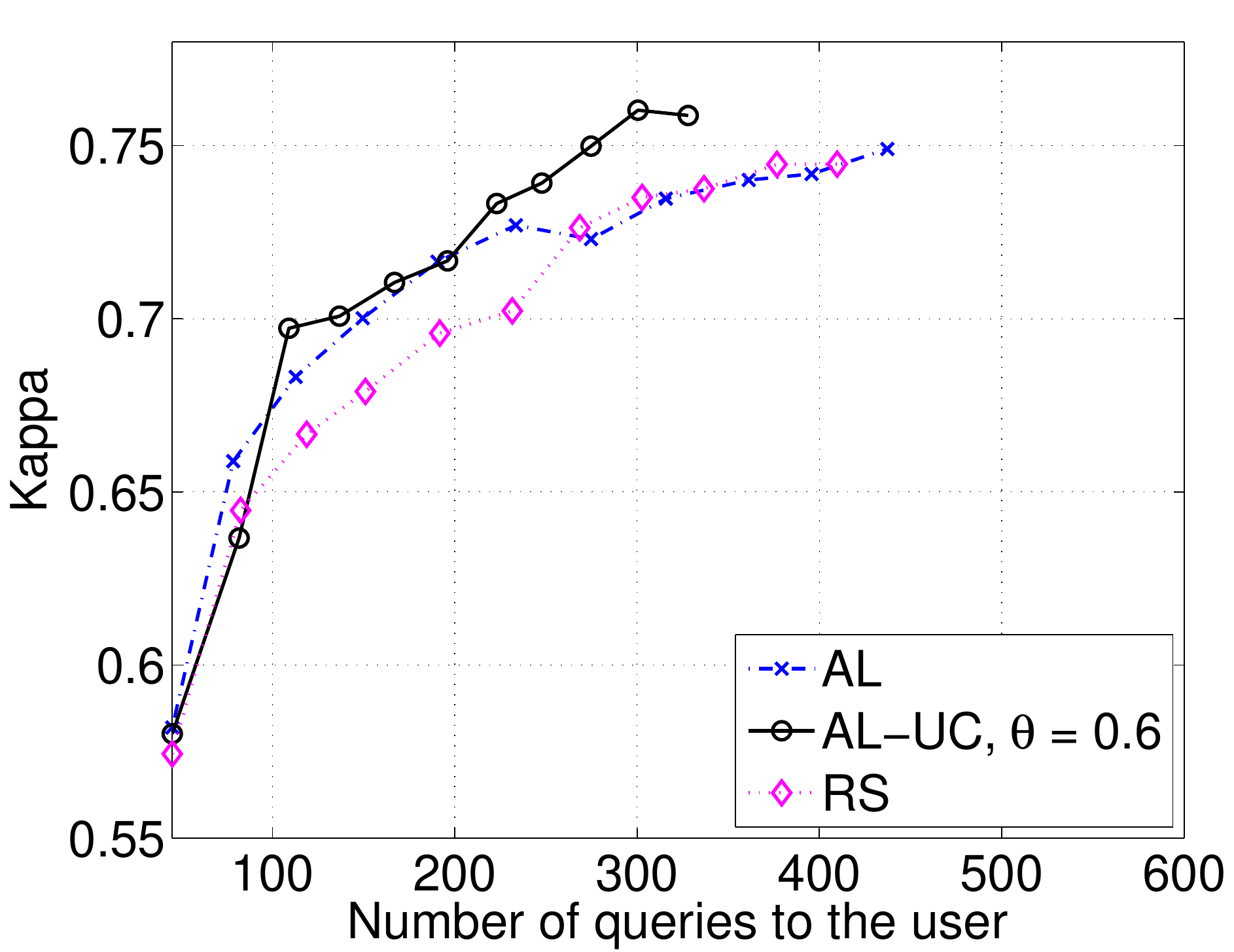}\\
& a) & b) & c)
\end{tabular}
\caption{Numeric results for the Br\"uttisellen dataset. a) Kappa statistic with traditional active learning setting; b) Number of queries per iteration involving 20 valid labeled samples; c) Kappa statistic related to average real effort provided by the user.
}
\label{fig:curves}
\end{figure*}

Finally, both sets of information are fused by creating a confidence mask. The confidence maps of Fig.~\ref{fig:maps}(b) are simply thresholded at a level of confidence $\theta$ and only the pixels for those $p((y_{\theta, i} = +1 | \x)) > \theta$ become presentable to the user. This way, the model continues to rank the pixels according to the active learning criterion, but only areas supposed to be easily understood by the user are made visible (in red in Fig.~\ref{fig:maps}(c)).

Numerical results on the Br\"uttisellen image are reported in Fig.~\ref{fig:curves} for the three methods considered \blue{and the two heuristics tested}.
The left-hand panel (Fig.~\ref{fig:curves}(a)) illustrates the numerical performance at the end of each iteration. At first glance, the proposed AL-UC seems not to outperform the standard AL.

These observations would be correct considering an omniscient user who can always label the pixels queried by the model (i.e. if $20$ labeled pixels could be obtained with $20$ queries per iteration). However, as illustrated in Fig.~\ref{fig:pix}, the traditional active learning method often highlights border pixels that the user cannot label: after initialization, where $45$ pixels are queried by each method, the user needs to consider, on average, $45$ to $50$ pixels to provide $20$ labels at every iteration (Fig.~\ref{fig:curves}(b)). Random sampling is much more efficient in this sense, since the user needs around $35$ queries to label $20$ pixels (40\% of the queries were feasible for the user). Finally, the proposed AL-UC was the most efficient, since only $25$ to $30$ queries are necessary to obtain the $20$ labels. This shows that the second classifier has  learned the confidence of the user, as it queries more useful pixels than the random strategy.

This leads to a rescaling of the curves of Fig.~\ref{fig:curves}(a) into a more realistic picture, that is illustrated in Fig.~\ref{fig:curves}(c): the performance is plotted as a function of the real effort provided by the user, i.e., the total number of queries (both successful and unsuccessful). This figure shows that the proposed method has a steeper learning phase than the traditional active learning method, since the latter wastes several queries on pixels too difficult to label. When taking a performance objective of 0.75 in $\kappa$, the proposed method needed on the average $432$ queries to reach it, while traditional active learning needed one hundred additional queries.

By looking back at the confidence maps in Fig.~\ref{fig:maps}(b), we can also observe that during the iterations the confidence is spread in all the areas that are easy to label: regular areas, like buildings, bare soil or very contrasted structures, such as roads. These maps also illustrate that the threshold of confidence $\theta$ must not be chosen too high, since a risk of getting stuck on a single object in the first iterations increase: in this case, setting $\theta = 0.9$ would have focused all the sampling in a few areas such as the commercial area in the bottom-left corner, thus trapping the solution in a local maximum.

\begin{figure}[!b]
\centering
\begin{tabular}{cc}
\includegraphics[width=4cm]{./figstgrs/brut-MCLU-2-cmr}&
\includegraphics[width=4cm]{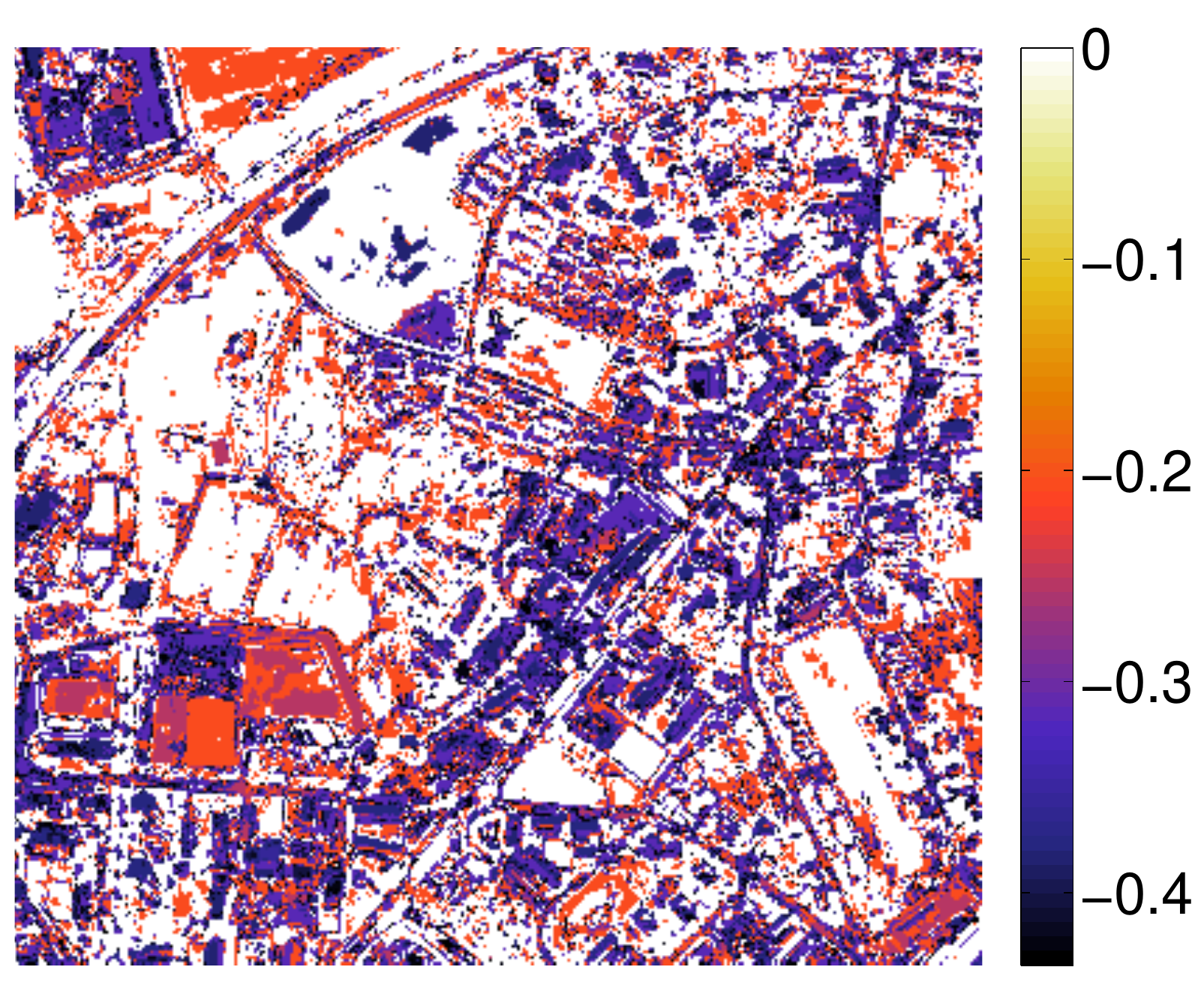}\\
(a) MCLU &(b) EQB\\
\end{tabular}
\caption{\blue{Comparison of the uncertainty maps returned by MCLU and EQB (multiplied by a -1 factor). Dark areas correspond to uncertain areas.}}
\label{fig:heu}
\end{figure}

\blue{Regarding comparison with the EQB heuristic (reported in the bottom row of Fig.~\ref{fig:curves}) similar trends are observed. The learning curve is less steep, showing the adequacy of MCLU for active learning with SVM (the heuristic uses the decision function directly, while EQB  works with committees of models), but in general both reach a similar performance after 200 valid queries. However, EQB seems to require less queries to find the 20 valid samples, at each iteration and independently from the approach tested. This can be explained by the differences in the nature of the two heuristics (Fig.~\ref{fig:heu}): MCLU ranks pixels according to the (continuous) decision function obtained on the two most probable classes; this means that there are potentially as many different values as candidates in $U^\epsilon$. We saw that, on the average, 50\% of the samples minimizing such function are easy to label. On the contrary, the EQB heuristic depends on a committee of predictions and the heuristic only has a limited number of possible entropy values (depending on the number of classes $\Omega$, the number of models $k$ and the number of classes predicted by the committee $N_i$ (see Eq.~\eqref{eq:neqb}). Therefore, the EQB function is a quantized function with few distinct values\footnote{\blue{In the case of the experiments reported, the number of partitions $P$ (corresponding to different entropy values) achievable by a committee of $n$ models predicting $K$ classes is $P(n,K) = \sum_{k=1}^{K} P(n,k)$. To compute this quantity, we use the following three properties: i) $P(n,k) = P(n-1,k-1) + P(n-k,k)$; ii) $P(n,k) = 0$ if $n < k$; iii) $P(n,n) = P(n,1) =  1$. Using the recursive formula in i), the maximum number of entropy values is $P(10,9) = \sum_{k = 1}^9 P(10,k) =  41$.}}. As a consequence, several pixel candidates receive the maximal entropy value and the choice is then done randomly among those with maximal entropy. Consequently with EQB, the chances to query a pixel that the user can label increase.}

\begin{figure}[!t]
\centering
\begin{tabular}{c}
\includegraphics[width=0.4\textwidth]{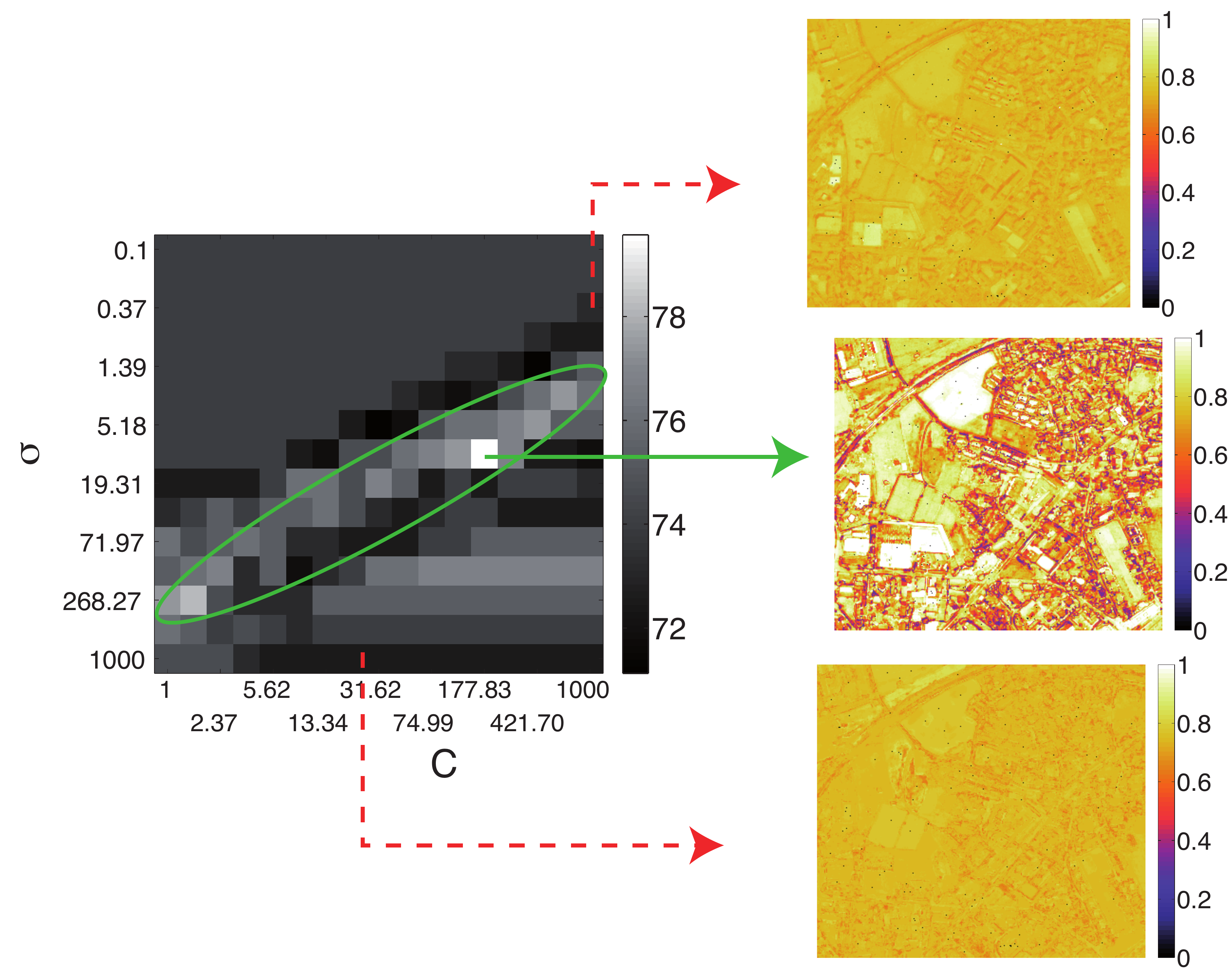}
\end{tabular}
\caption{Crossvalidation surface showing the overall accuracy of the confidence classifier for the Br\"uttisellen dataset. On the right, confidence maps corresponding to the maximum (in green/solid) and two minima (in red/dashed).}
\label{fig:surf}
\end{figure}

Figure~\ref{fig:surf} illustrates the importance of finding a good model to describe the confidence. It shows the crossvalidation surface of the parameter space of the confidence classifier at iteration $\epsilon = 4$ for a given run of the AL-UC algorithm. The green circle highlights the area of solutions capable of describing the confidence of the user. A good selection of parameters results in confidence maps that successfully constrain the AL heuristic (in green/solid), whereas a bad choice of parameters leads to confidence maps that do not constrain the AL heuristic (in red/dashed).

\subsection{Zurich highway}\label{sec:High}

\begin{figure*}[!t]
\centering
\begin{tabular}{ccc}
\includegraphics[width=5.3cm]{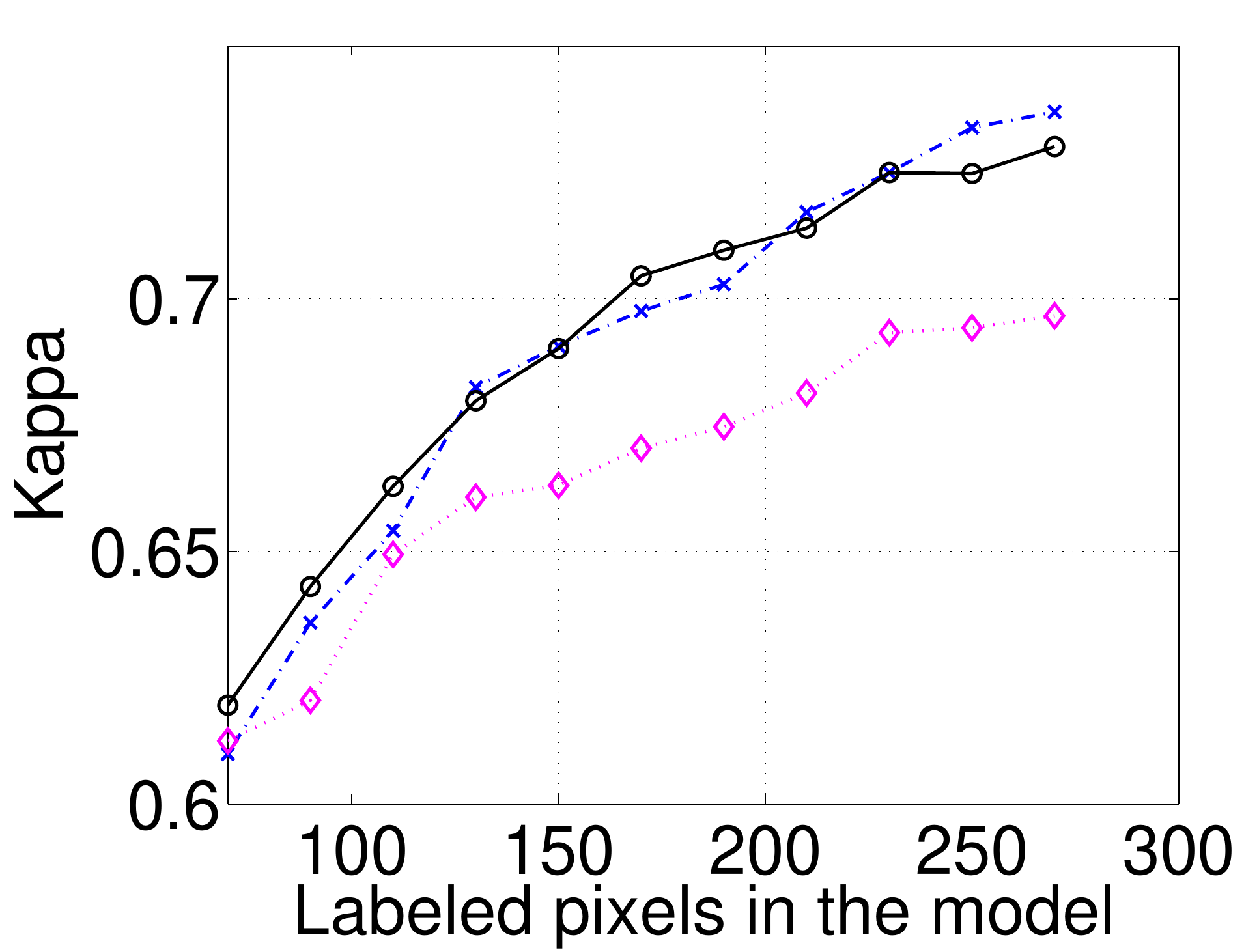}&
\includegraphics[width=5.3cm]{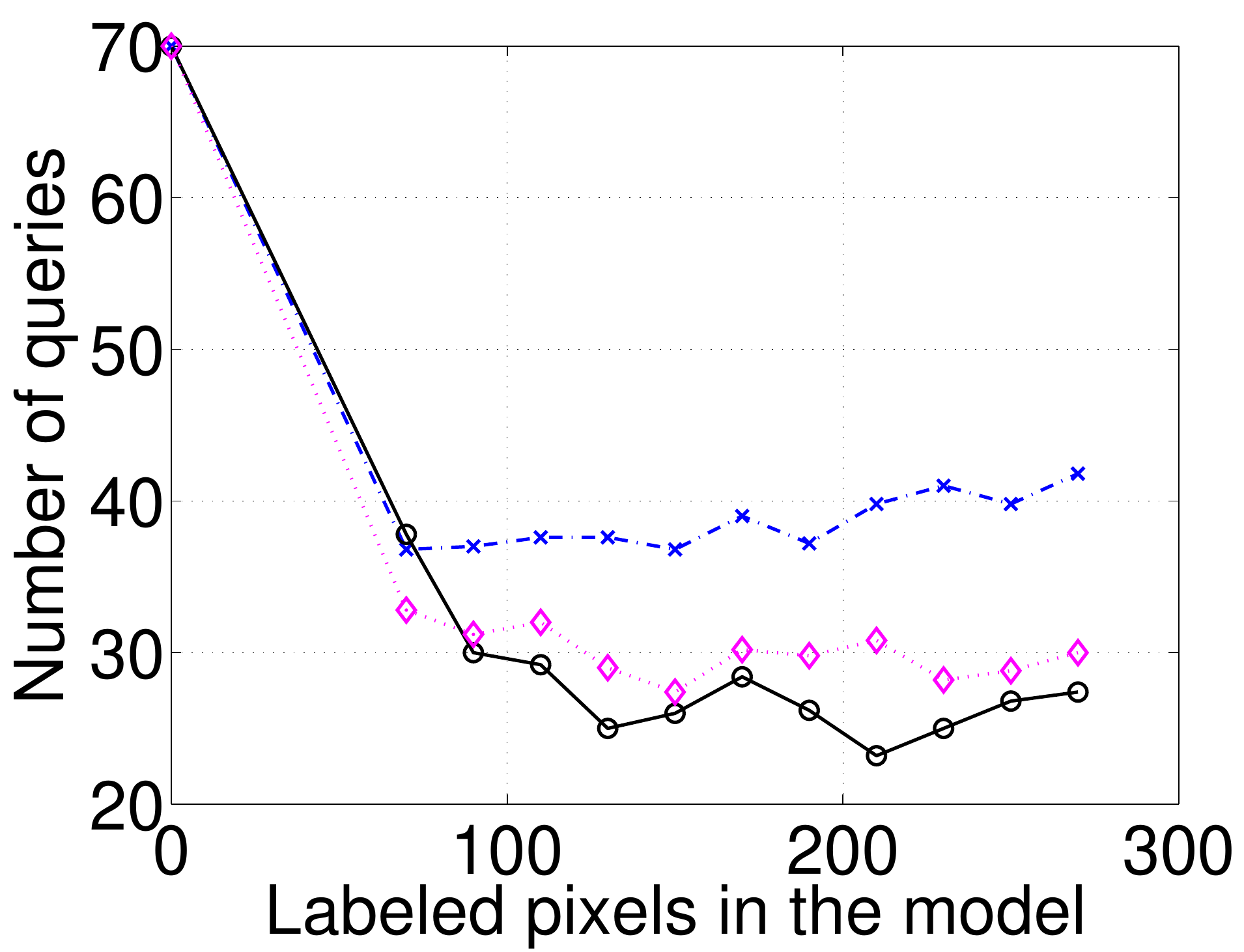}&
\includegraphics[width=5.3cm]{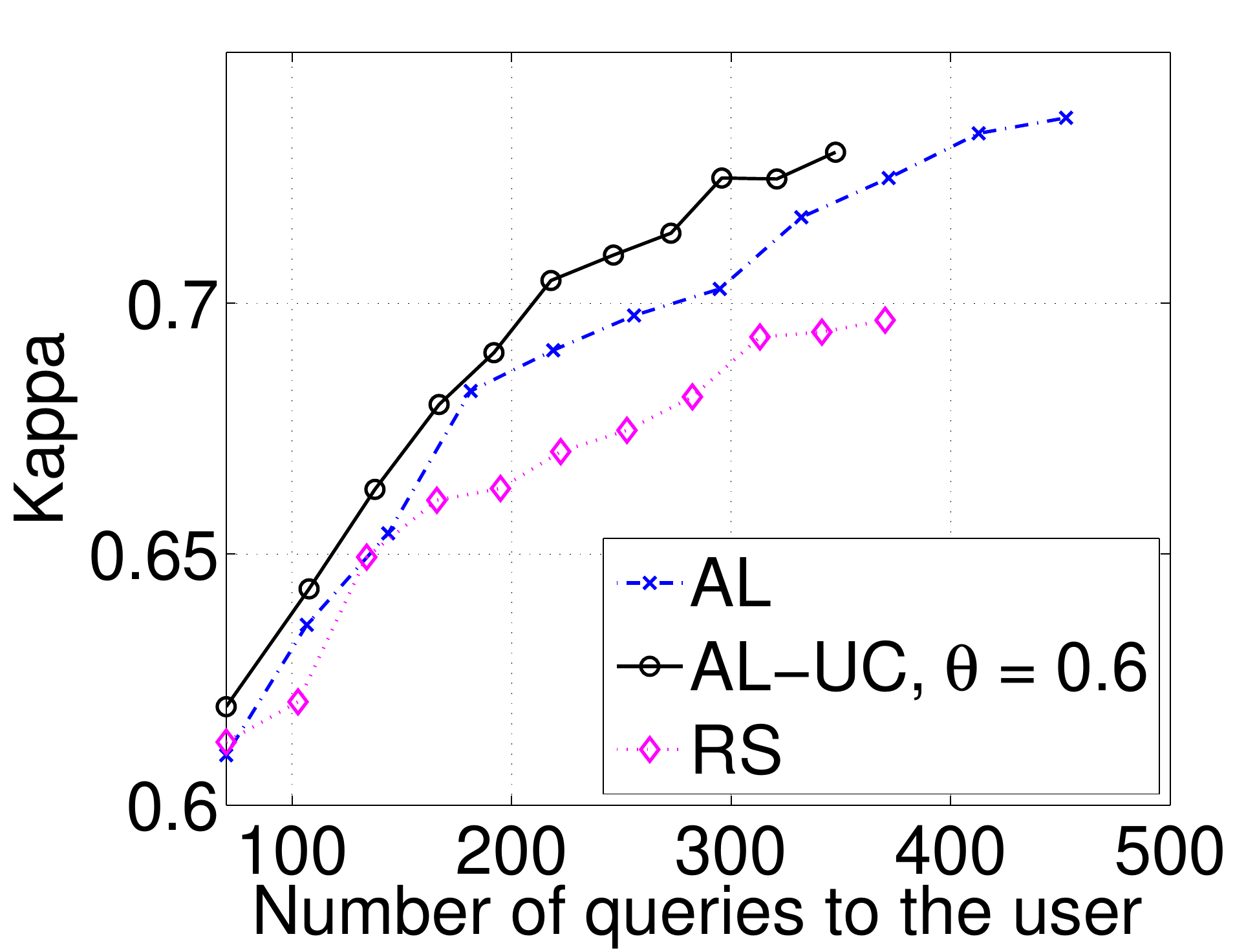}\\
a) & b) & c)
\end{tabular}
\caption{Numeric results for the Highway dataset using the MCLU heuristic. a) Kappa statistic with traditional active learning setting; b) Number of queries per iteration involving 20 valid labeled samples; c) Kappa statistic related to average real effort provided by the user.}
\label{fig:curvesH}
\end{figure*}

\begin{figure}[!t]
\centering
\begin{tabular}{cc}
\includegraphics[height=3.7cm]{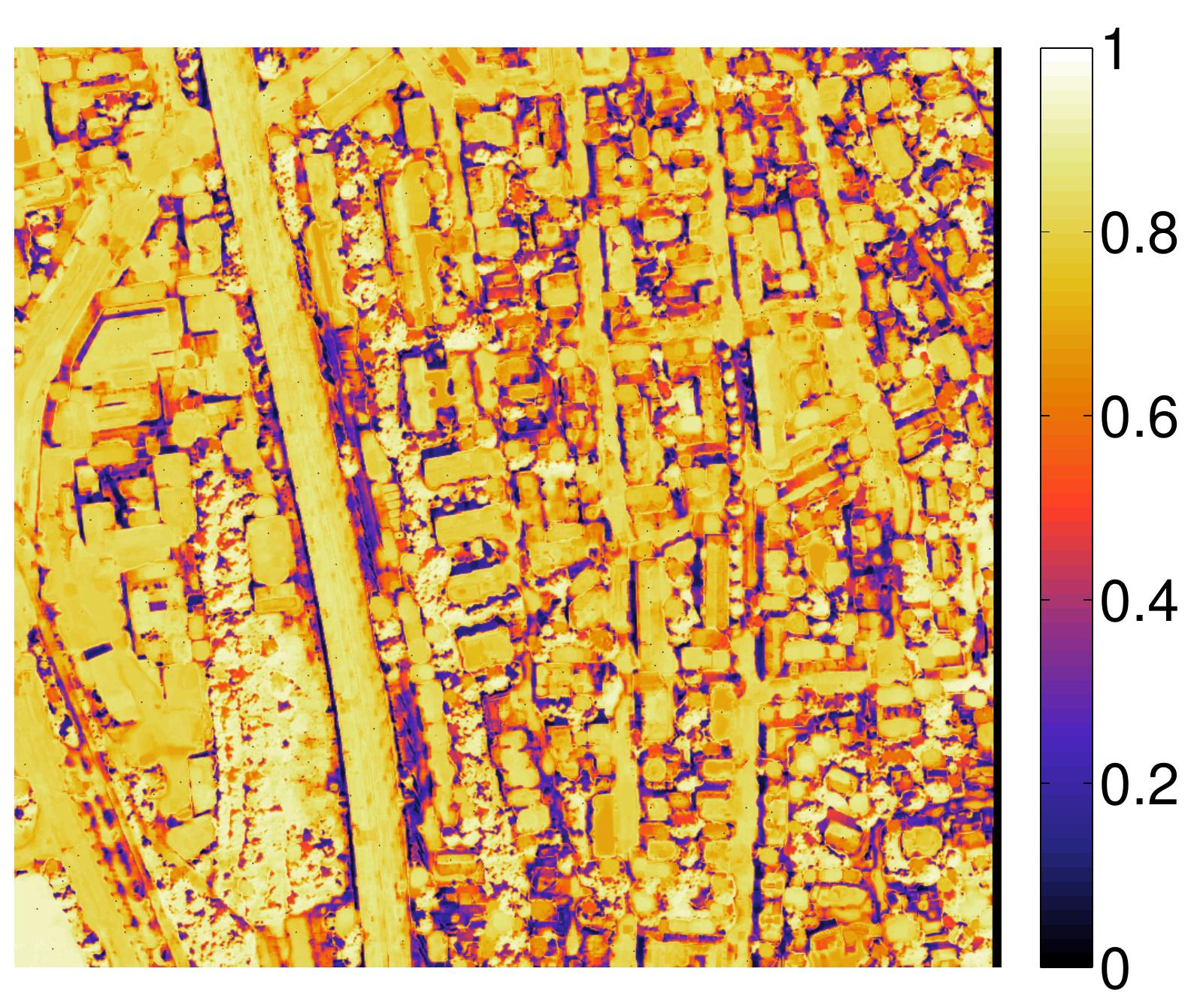}&
\includegraphics[height=3.8cm]{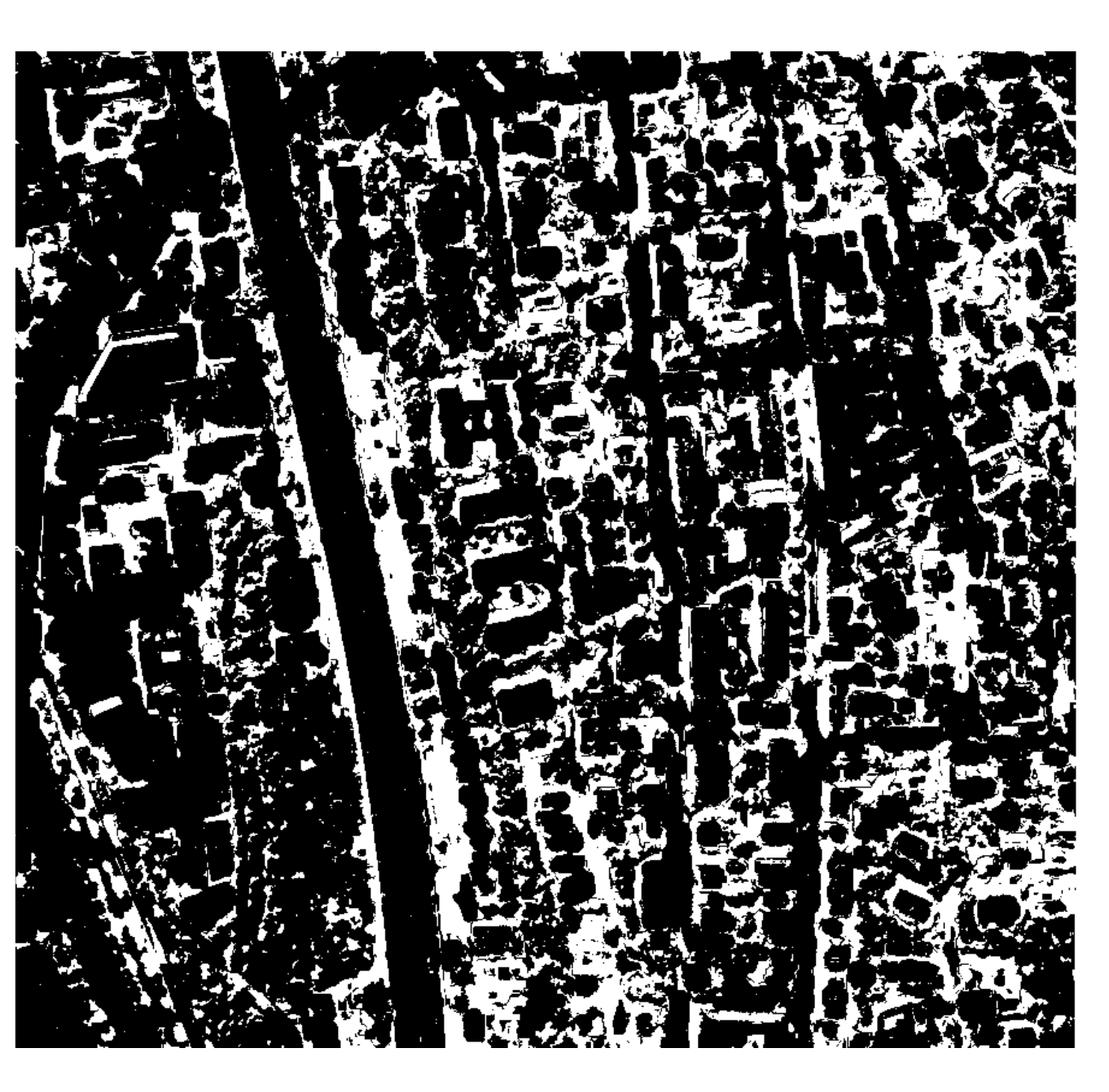}\\
(a) Confidence map & (b) Confidence mask\\
\end{tabular}
\caption{(a) Confidence map and (b) corresponding binary mask for the Highway dataset at iteration $\epsilon = 5, |X^5| = 170, |X_\theta^5| = 218$. }
\label{fig:confHigh}
\end{figure}

Experimental results on the Highway dataset are reported in Fig.~\ref{fig:curvesH} for the three methods considered. Given the 	higher resolution, the difference between the compared methods is expected to be lower, since a higher proportion of uncertain samples are expected to be within the objects of interest (for instance chimneys on roofs). This is due to the increased intraclass variance of the classes in the Highway dataset, that has a spatial resolution four times higher ($0.6$ m for ``Highway'' vs. $2.4$ m for ``Br\"uttisellen''). As a consequence, the ambiguous areas are more limited in space (Fig.~\ref{fig:confHigh}), since the traditional AL tends to ask pixels that are anomalous, but within the urban objects: in this case, the user can respond more easily to the query, since the object itself is easily identifiable. 

Curves reported in Fig.~\ref{fig:curvesH} confirm this intuition: after initialization, where $70$ pixels are queried for each method, AL-UC performs similarly to classical active learning, but AL requires on average $10$ additional queries per iteration to obtain the $20$ labeled pixels. On  average, AL-UC needs $25$ to $30$ queries (similar to RS), while AL requires about $40$. The right-hand plot illustrates the effort demanded to the used against the performance and, as for the previous dataset, AL-UC allows to retrieve higher accuracy with less queries.

\subsection{\blue{Considering a committee of users}}\label{sec:users}
\begin{figure}[!t]
\centering
\includegraphics[height=4.7cm]{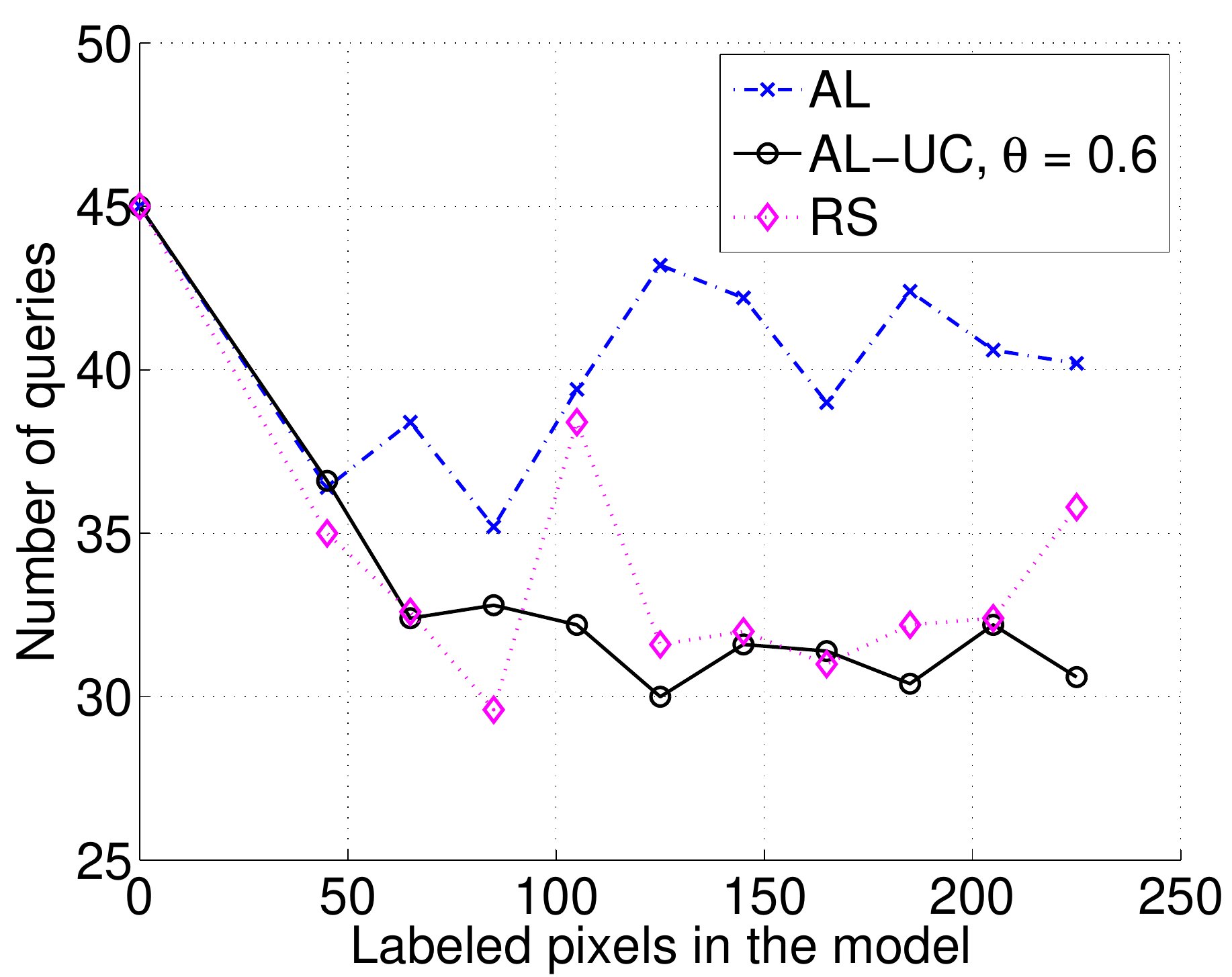}
\caption{Average number of queries for a committee of different users performing a single run of each model.}
\label{fig:users}
\end{figure}

\blue{In this section, we consider a committee of five users, each one performing one experiment on the Brutisellen image with the three models. Three of them are trained remote sensing analysts, while the  others two are not familiar with labeling tasks. Figure~\ref{fig:users} illustrates the average number of queries required by the four users. The tendency observed in the single-user experiment presented above are confirmed. This shows that the proposed method efficiently evaluates the confidence of the user in labeling, avoiding bad states and significantly reducing the number of queries.}

\section{Conclusions}
\label{sec:concl}

In this paper, we studied a way to learn the confidence of a photointerpreter to make active learning routines effective in real-life scenarios. For the first time, an active learning method is assessed on the entirety of an image and shortcomings related to the uncertainty of a signal are put in relation with the capacity of a user to provide a reliable label for the pixel. By assessing the probability of having a bad state (a pixel that the photointerpreter is unable to label), the most uncertain pixels ranked by active learning are filtered, thus presenting to the user a set that he/she is capable of labeling. Experiments on two QuickBird images at different spatial resolutions showed the efficiency of the method, which significantly decreased the number of queries that the user must provide to fulfill his  task.

This work opens a wide range of possible studies for operative active learning: the effects of the user  must be studied, as well as the effect of adapting the threshold $\theta$ along the iterations (see \cite{Che09}). If the latter is more a technical question to be tackled in the future, the first opens avenues related to crowdsourcing~\cite{Abe11,Gom11} and community-based online learning of surface signatures.

\section*{Acknowledgements}
The authors would like to acknowledge M. Kanevski (University of Lausanne) for the access to the QuickBird images and M. Volpi (University of Lausanne) for the important inputs about the paper. \blue{We also would like to thank the five users for collaborating in the users committee experiments.}

\bibliographystyle{IEEEbib}
\bibliography{al-rw}

\end{document}